\documentclass[letterpaper]{article} 
\usepackage{aaai25}  
\usepackage{times}  
\usepackage{helvet}  
\usepackage{courier}  
\usepackage[hyphens]{url}  
\usepackage{graphicx} 
\def\UrlFont{\rm}  
\usepackage{natbib}  
\usepackage{caption} 

\usepackage{subfiles} %
\usepackage{multirow} %
\usepackage{makecell}
\usepackage{colortbl}
\usepackage{pifont}%
\usepackage[switch]{lineno}
\usepackage[misc]{ifsym}
\usepackage{bm}
\usepackage{anyfontsize}
\usepackage{algorithm}
\usepackage{algorithmic}

\usepackage{newfloat}
\usepackage{listings}
\DeclareCaptionStyle{ruled}{labelfont=normalfont,labelsep=colon,strut=off} 
\floatstyle{ruled}
\newfloat{listing}{tb}{lst}{}
\floatname{listing}{Listing}
\usepackage{mathtools} 
\usepackage{graphics}
\usepackage{subcaption}
\usepackage{bbding}
\usepackage{booktabs}
\usepackage{enumitem}
\usepackage{amsmath}
\usepackage{amssymb}
\usepackage[dvipsnames]{xcolor}
\definecolor{dt}{gray}{0.7}  

\definecolor{mycolor}{HTML}{D8ECD1}

\title{CLIP-CID: Efficient CLIP Distillation via Cluster-Instance Discrimination}
\author{
    Kaicheng Yang\textsuperscript{\rm 1}\equalcontrib,
    Tiancheng Gu\textsuperscript{\rm 2}\equalcontrib,
    Xiang An\textsuperscript{\rm 1},
    Haiqiang Jiang\textsuperscript{\rm 1}, \\
    Xiangzi Dai\textsuperscript{\rm 1}, 
    Ziyong Feng\textsuperscript{\rm 1},
    Weidong Cai\textsuperscript{\rm 2}$^{\ddagger}$,
    Jiankang Deng\textsuperscript{\rm 3}$^{\ddagger}$\\
}
\affiliations{
    \textsuperscript{\rm 1} DeepGlint \\
    \textsuperscript{\rm 2} University of Sydney \\
    \textsuperscript{\rm 3} Imperial College \\
    \texttt\small{kaichengyang@deepglint.com}
}

\begin{document}

\maketitle

\begin{abstract}
Contrastive Language-Image Pre-training (CLIP) has achieved excellent performance over a wide range of tasks. However, the effectiveness of CLIP heavily relies on a substantial corpus of pre-training data, resulting in notable consumption of computational resources. Although knowledge distillation has been widely applied in single modality models, how to efficiently expand knowledge distillation to vision-language foundation models with extensive data remains relatively unexplored. In this paper, we introduce CLIP-CID, a novel distillation mechanism that effectively transfers knowledge from a large vision-language foundation model to a smaller model. We initially propose a simple but efficient image semantic balance method to reduce transfer learning bias and improve distillation efficiency. This method filters out 43.7\% of image-text pairs from the LAION400M while maintaining superior performance. After that, we leverage cluster-instance discrimination to facilitate knowledge transfer from the teacher model to the student model, thereby empowering the student model to acquire a holistic semantic comprehension of the pre-training data. Experimental results demonstrate that CLIP-CID achieves state-of-the-art performance on various downstream tasks including linear probe and zero-shot classification.

\end{abstract}

\section{Introduction}
With the proliferation of mobile networks and social platforms, there has been an explosion in the production of image-text pairs~\cite{guo2019deep,gu2024rwkvclip}. This abundance of data has provided a strong foundation for the advancement of vision-language pre-training~\cite{radford2021learning,jia2021scaling,yu2020ch,yang2020cm}. The Contrastive Language-Image Pre-training (CLIP)~\cite{radford2021learning} achieves remarkable success in multi-modal learning by aligning image-text pairs on a large-scale dataset. It learns two separate unimodal encoders for image and text using a contrastive loss, one of the most effective losses for representation learning~\cite{tian2020contrastive,he2020momentum,chen2020simple,chopra2005learning}. Nevertheless, the efficacy of CLIP heavily depends on an extensive pre-training dataset. The original CLIP models are pre-trained on 400 million image-text pairs for 32 epochs, demanding thousands of GPU days. This poses a substantial challenge in scenarios with limited computational resources~\cite{radenovic2023filtering,yang2023alip}. Recently, large-scale image-text datasets crawled from websites, such as LAION400M~\cite{schuhmann2021laion} and LAION5B~\cite{schuhmann2022laion}, have gained widespread usage in vision-language pre-taining. DataComp~\cite{gadre2024datacomp} consists of image-text pairs extracted from Common Crawl's web data and employs various strategies such as basic filtering, CLIP score filtering, and text\&image-based filtering. However, there is still a lot of semantic repetition in the training data, which not only has the potential to impact the performance of representation learning but also results in a waste of computational resources~\cite{radenovic2023filtering,wang2023too}.

\begin{figure}[t!]
  \centering
  \includegraphics[width=\linewidth]{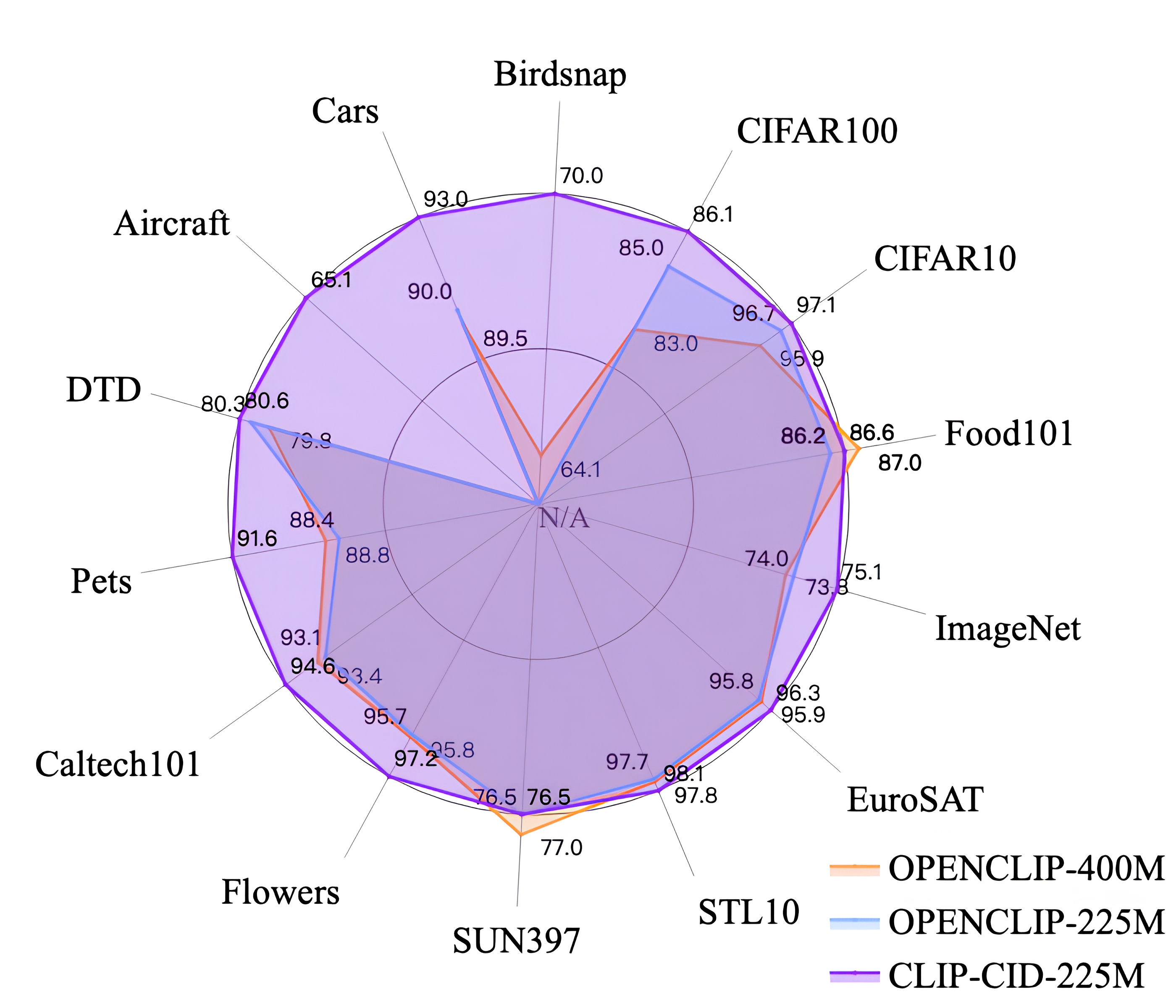}
  \caption{The linear probe performance comparison between CLIP-CID and OPENCLIP across 14 common datasets. Despite the exclusion of 43.7\% of image-text pairs from the LAION400M, CLIP-CID exhibits exceptional performance.}
  \label{fig:rader_graph}
\end{figure}

Knowledge Distillation (KD)~\cite{hinton2015distilling} is proposed to enhance the performance of a small student model by transferring knowledge from a large teacher model throughout the training phase. Most existing KD approaches in the literature are primarily designed for small-scale datasets ($e.g.$, CIFAR10 and CIFAR100~\cite{krizhevsky2009learning}), as well as small models ($e.g.$, ResNet50 and ResNet34~\cite{he2016deep}). Recent studies have concentrated on distilling CLIP for particular target tasks or datasets. For example, BeamCLIP~\cite{kim2022transferring} introduces cross-modal similarity matching and context-based prompt augmentation to transfer knowledge from CLIP representations to a small model, achieving better performance on the ImageNet~\cite{deng2009imagenet}. ZeroSeg~\cite{chen2023exploring} distills the visual concepts learned by CLIP into a set of segment tokens, leading to a marked enhancement in training efficiency while preserving segmentation performance. Nevertheless, the exploration of leveraging knowledge distillation to enhance foundational models remains relatively sparse. Recent works~\cite{wu2023tinyclip,sun2023dime} successfully transfer knowledge from a large foundation model to a small one. However, their attention is restricted to instance-level knowledge, thus unable to capture the semantic structure of extensive training data effectively. This limitation arises from the nature of instance-wise contrastive learning, which consistently regards samples from distinct instances as negative pairs, disregarding their semantic similarity.

To address the aforementioned challenges, this paper introduces CLIP-CID, a novel distillation mechanism that effectively transfers knowledge from a large vision-language foundation model to a smaller model. In order to reduce transfer learning bias and improve distillation efficiency, we initially propose a simple but efficient method to balance the semantic concepts within the LAION400M dataset, which can filter out 43.7\% of the training data while maintaining superior performance. Furthermore, we enhance the knowledge transfer from a large teacher model to a smaller student model by integrating cluster-instance discrimination, which facilitates a more comprehensive semantic understanding of the student model. As illustrated in Fig.~\ref{fig:rader_graph}, our proposed CLIP-CID demonstrates superior linear probe performance across 14 common datasets even after filtering out 43.7\% of image-text pairs from the LAION400M. The main contributions of this paper are summarized as follows:

\begin{itemize}
\item We propose a simple but efficient image semantic balance method to reduce transfer learning bias and improve distillation efficiency, which can remove 43.7\% of image-text pairs from the LAION400M while maintaining superior performance.

\item We introduce the CLIP-CID, a novel distillation mechanism that integrates cluster discrimination and instance discrimination to effectively transfer knowledge from a large vision-language foundation model to a smaller model.

\item We conduct extensive experiments to validate the effectiveness of our proposed approach. The experimental results prove that CLIP-CID achieves state-of-the-art performance on various downstream tasks, including zero-shot classification and linear probe.
\end{itemize}

\section{Related Work}

\subsection{Vision-Language Pre-training}
CLIP~\cite{radford2021learning} has achieved remarkable attention due to its exceptional zero-shot recognition ability and successful transfer capabilities. Recent developments have improved CLIP-based methodologies~\cite{mu2022slip,geng2023hiclip}. ALBEF~\cite{li2021align} introduces a contrastive loss to align the image and text representations before fusing them through cross-modal attention, which enables more grounded vision and language representation learning. ALIGN~\cite{jia2021scaling} leverages a dataset of over one billion noisy image alt-text pairs to scale visual and vision-language representation learning. FILIP~\cite{yao2021filip} successfully exploits the finer-grained expressiveness between image patches and textual words by modifying the contrastive loss and enables offline pre-computation of image and text representations during inference. FLIP~\cite{li2023scaling} employs random masking and removal of a significant portion of image patches during training, enabling learning from a larger number of image-text pairs within the same wall-clock time and contrasting more samples per iteration while maintaining a comparable memory footprint. However, pre-training vision-language foundation models on extensive datasets present a substantial challenge due to the high costs involved and the significant consumption of computational resources.

\subsection{Large-Scale Dataset Filtering}
While open-source large-scale datasets such as LAION400M~\cite{schuhmann2021laion} have filtered image-text pairs based on CLIP scores below 0.3, it is crucial to emphasize that the effectiveness of CLIP scores may be compromised in specific scenarios. For instance, images containing visible text within the image often yield high CLIP scores but could prove detrimental to representation learning~\cite{cao2023less}. Recent research proposes a complexity, action, and textspotting filtering strategy~\cite{radenovic2023filtering} to select informative image-text pairs from noisy web-scale datasets. Additionally, the pruning method adapted from ImageNet to LAION~\cite{abbas2024effective} uses a straightforward complexity measure to reduce training costs to one-quarter of the standard approach. However, the process of cleaning large-scale datasets presents a computational challenge. To tackle this issue, SemDeDup~\cite{abbas2023semdedup} leverages embeddings from pre-trained models to identify and eliminate semantic duplicates. Different from the above methods, we propose a simple but efficient image semantic balance method to reduce transfer learning bias and improve distillation efficiency. It only entails a single traversal of the data and can filter 43.7\% of image-text pairs from the LAION400M while maintaining superior performance.

\subsection{Knowledge Distillation}
Knowledge distillation has found wide application in the field of computer vision. Building upon the achievements of visual-language foundational models, recent research efforts~\cite{wei2022contrastive,dong2022bootstrapped} have demonstrated significant performance enhancements on specific datasets, such as ImageNet~\cite{deng2009imagenet}. Hybrid Distillation~\cite{shi2023hybrid} achieves superior performance on COCO~\cite{lin2014microsoft} by integrating masked autoencoders with CLIP. CLIP-TD~\cite{wang2022clip} efficiently distills knowledge from CLIP into established architectures through a dynamically weighted objective concentrating on adaptively chosen tokens per instance, resulting in remarkable performance gains on visual commonsense reasoning tasks. DIME-FM~\cite{sun2023dime} aims to distillate small foundation models using smaller-scale public images and unpaired sentences. However, the above methods only focus on instance-level knowledge, neglecting the semantic structure of large-scale training data. This limitation arises from instance-wise contrastive learning, which pairs samples as negatives based solely on their instance differences, without considering their semantic similarity.

\section{Methodology}
\subsection{Image Semantic Balance} 
\label{sec:image_balance}

\begin{figure}[!t]
\centering
    \begin{subfigure}{0.23\textwidth}
    \includegraphics[width=1.0\textwidth]{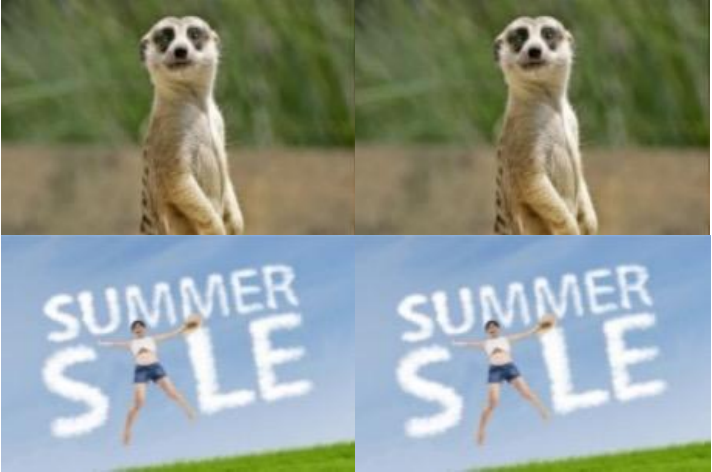}
    \caption{\small Perceptual Redundancy}
    \label{fig:perceptual_duplication}
    \end{subfigure}
    \begin{subfigure}{0.23\textwidth}
    \includegraphics[width=1.0\textwidth]{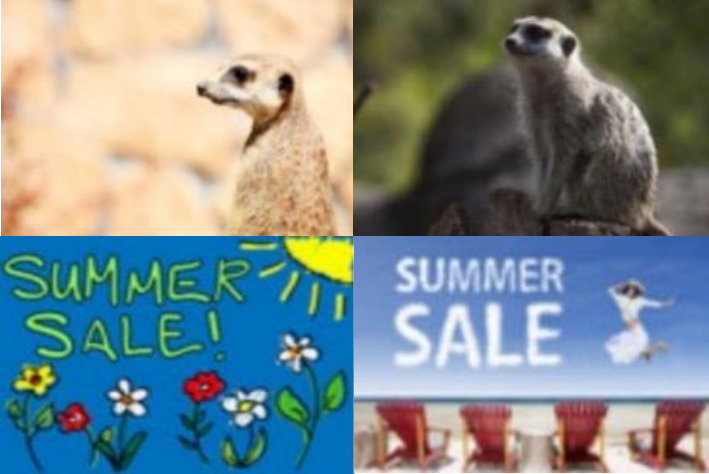}
    \caption{\small Semantic Redundancy}
    \label{fig:semantic_duplication}
    \end{subfigure}
    \begin{subfigure}{\linewidth}
        \centering
        \includegraphics[width=0.98\linewidth]{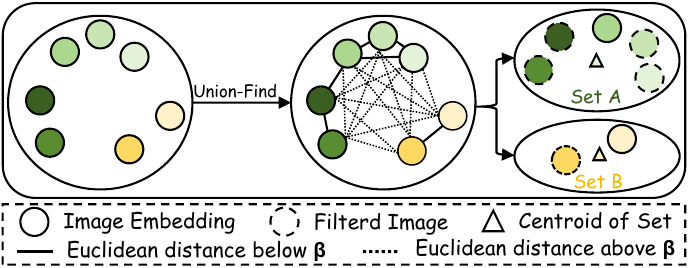}
        \caption{Visualization of the image semantic balance process.}
        \label{fig:balance_process}
  \end{subfigure}
  \begin{subfigure}{\linewidth}
        \centering
        \includegraphics[width=\linewidth]{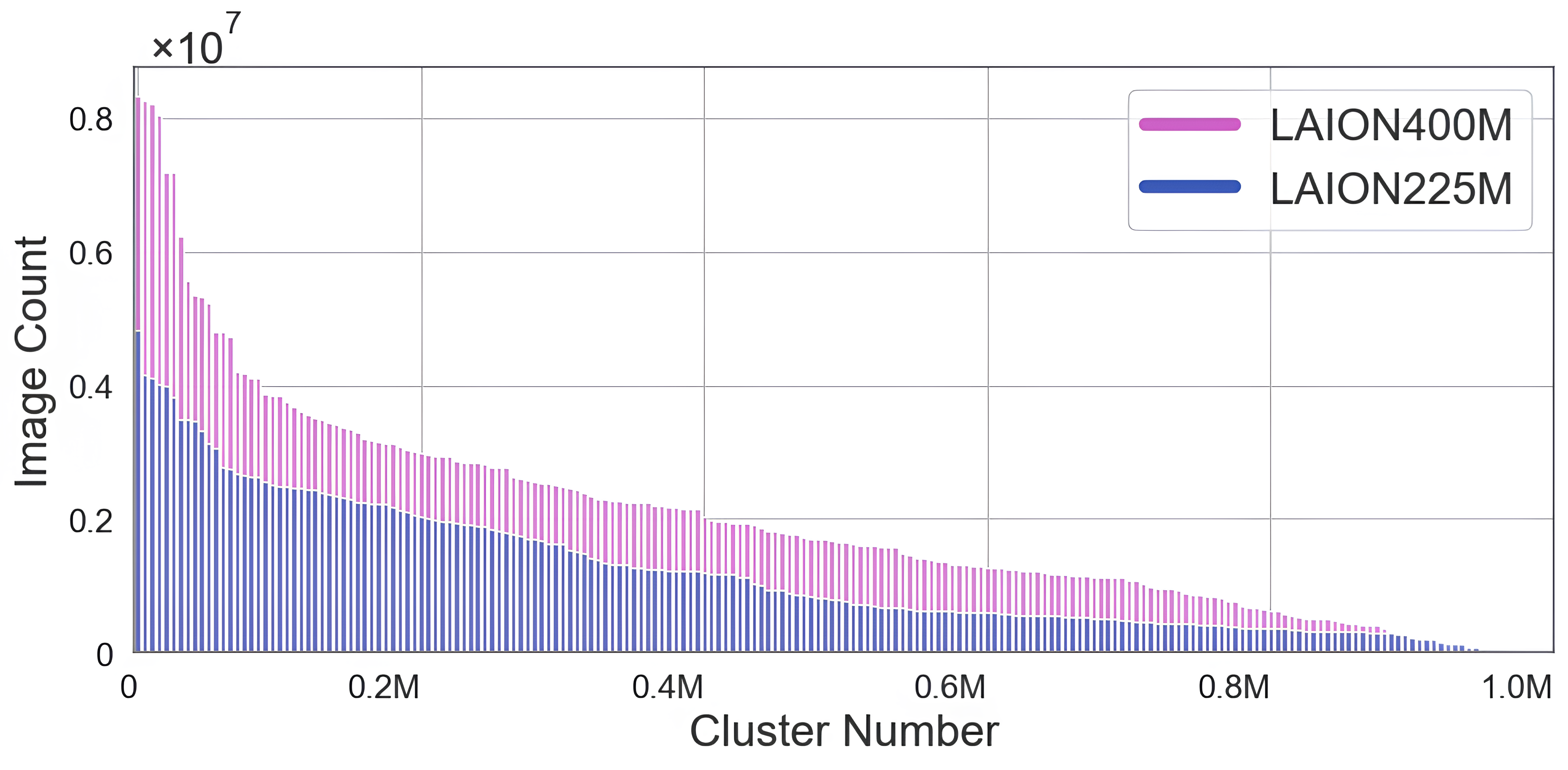}
        \caption{Distribution of LAION400M and LAION225M.}
        \label{fig:balance}
  \end{subfigure}

\caption{(a) and (b) visualization of the perceptual redundancy images and semantic redundancy images. (c) visualization of the image semantic balance process. (d) distribution of LAION400M and LAION225M in 1M clusters.}
\vspace{-3mm}
\end{figure}

To enhance the comprehensive transfer of knowledge from the teacher model to the student model and reduce transfer learning bias, we propose a simple yet effective image semantic balance method. Our primary objective is to address both perceptual redundancy and semantic redundancy within images. As illustrated in Fig.~\ref{fig:perceptual_duplication}, perceptual redundancy images showcase minimal discrepancies at the pixel level. In contrast, as depicted in Fig.~\ref{fig:semantic_duplication}, semantic redundancy images exhibit significant pixel-level variations while maintaining notably similar semantic information. As shown in Fig.~\ref{fig:balance}, the presence of perceptual redundancy and semantic redundancy in images causes an imbalanced distribution of semantic concepts in the dataset, leading to knowledge transfer biases in the student model during distillation.

In contrast to existing cluster-based methods that require multiple iterations~\cite{abbas2024effective,abbas2023semdedup}, our method requires only a single complete traversal of the training data to integrate images with similar semantics into the same set. The image semantic balance process is shown in Fig.~\ref{fig:balance_process}. Initially, we employ the OPENCLIP ViT-bigG/14~\cite{ilharco_gabriel_2021_5143773} model to extract image embeddings from LAION400M~\cite{schuhmann2021laion}. To address memory limitations, we divide all image embeddings $E\in \mathbb{R}^{N \times d}$ into $c$ chunks and distribute $E_{n}\in \mathbb{R}^{\frac{N}{c} \times d}$ to different nodes. Then we calculate the Euclidean distance~\cite{dokmanic2015euclidean} matrix between the current image and the images contained in various chunks, arranging the rows in ascending order according to their distances. The top-$k$ results for each chunk are retained in the matrix $C_{n} \in \mathbb{R}^{\frac{N}{c} \times k}$, and these distance matrices are concatenated to form the global matrix $C \in \mathbb{R}^{N \times k}$. After that, we employ the Union-Find algorithm~\cite{tarjan1975efficiency} to group semantically similar images together. If the distance between different images surpasses the distance threshold $\beta$, they will be assigned to separate sets. Otherwise, they will be merged into the same set. We identify the central image (the image nearest to the centroid of the set) and eliminate the remaining images within the same set. Consequently, the total count of LAION400M images is decreased to 225M, denoted as LAION225M. It is worth noting that the GPU is exclusively used for computing embedding distances, while the union-find algorithm is executed on the CPU. As shown in Fig.~\ref{fig:balance}, LAION225M demonstrates a smoother distribution, facilitating the student model learning a more comprehensive knowledge from the teacher model.

\subsection{Cluster-level Distillation}
Traditional instance-wise contrastive learning treats different instances as negative pairs, limiting its ability to capture the complete semantic information in the training data~\cite{caron2018deep,asano2019self,zhan2020online,qian2022unsupervised}. In this study, we introduce cluster discrimination knowledge distillation to delve into potential semantic structures inherent in the training dataset. Our method involves grouping visually similar instances into clusters, thus enabling a more comprehensive semantic representation. The cluster discrimination distillation method involves two stages: (1)~Clustering, which assigns a unique class label to each image and gets cluster centers. (2)~Cluster discrimination distillation, which facilitates the transfer of global knowledge from the teacher model to the student model.

\paragraph{Clustering.} We investigate the standard $k$-means algorithm, which aims to partition a given set of vectors into $k$ distinct groups based on the nearest neighbor criterion. Given the normalized image embedding $e_{i}$, the clustering process involves jointly learning a centroid matrix $W \in \mathbb{R}^{d\times k}$ and assigning the cluster label $z_{i}$ for each image by solving the following optimization problem:

\begin{equation}
\small
\begin{aligned}
\min_{W\in \mathbb{R}^{d\times k}}
\frac{1}{N}
\sum_{i=1}^N
\min_{z_i \in \{0,1\}^{k}}
\| e_i  -  W z_i \|_2^2
\quad
\text{s.t.}
\quad
z_i^\top \bf{1}_k = 1,
\end{aligned}
\end{equation} where $N$ is the number of training samples, and the centroid $w_i$ belonging to centroid matrix $W\in \mathbb{R}^{d\times k}$ is considered the normalized prototype of $i$-th cluster. $z_i$ in  $\{0,1\}^k$ is a single label assignment restricted by the condition $z_i^\top \bf{1}_k = 1$, where $ \bf{1}_k $ is 1-vector with size of $k$. 

In this work, we employ the OPENCLIP ViT-bigG/14~\cite{ilharco_gabriel_2021_5143773} model to extract image embeddings. The automatically clustered large-scale dataset inevitably faces challenges such as intra-class impurity and inter-class conflicts due to the presence of noise in the large uncurated web-scale datasets. The intra-class impurity can be addressed by adjusting the cluster number. Meanwhile, the inter-class conflict can be effectively mitigated by reducing the number of sampled negative instances within the minibatch. Following previous works~\cite{unicom,mlcd}, we leverage the benefits of efficient feature quantization~\cite{johnson2019billion} to cluster LAION225M into one million classes. To alleviate inter-class conflict, we employ PatialFC~\cite{an2022killing} and randomly sample a portion of the negative class centers during each iteration.

\begin{figure}[t]
  \centering
  \includegraphics[width=\linewidth]{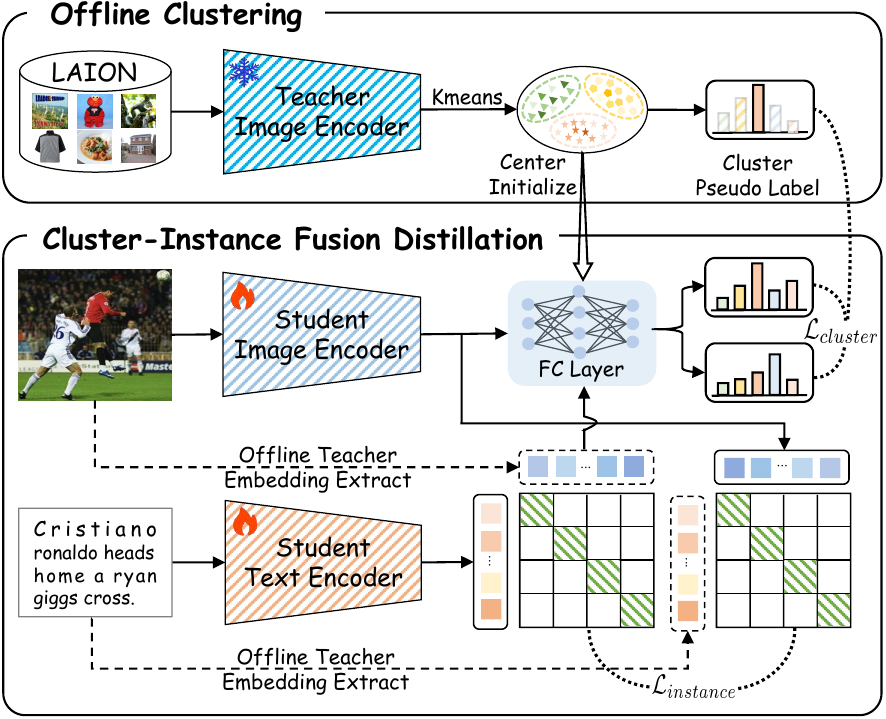} 
  \caption{
  The architecture of our proposed cluster-instance discrimination distillation.
  }
  \vspace{-3mm}
  \label{fig:architecture}
\end{figure}

\begin{table*}[!ht]
\begin{center}
\resizebox{\textwidth}{!}{
\begin{tabular}{l@{\hspace{1.1em}}c@{\hspace{0.5em}}c@{\hspace{0.6em}}c@{\hspace{0.6em}}c@{\hspace{0.6em}}c@{\hspace{0.6em}}c@{\hspace{0.6em}}c@{\hspace{0.6em}}c@{\hspace{0.6em}}c@{\hspace{0.6em}}c@{\hspace{0.6em}}c@{\hspace{0.6em}}c@{\hspace{0.6em}}c@{\hspace{0.6em}}c@{\hspace{0.6em}}c@{\hspace{0.6em}}c@{\hspace{0.6em}}}
\toprule
Model& Dataset& \rotatebox[origin=lb]{90}{\smash{\small Food101}} & \rotatebox[origin=lb]{90}{\smash{\small CIFAR10}} & \rotatebox[origin=lb]{90}{\smash{\small CIFAR100}} & \rotatebox[origin=lb]{90}{\smash{\small Birdsnap}} & \rotatebox[origin=lb]{90}{\smash{\small  Cars}} & \rotatebox[origin=lb]{90}{\smash{\small Aircraft}} & \rotatebox[origin=lb]{90}{\smash{\small DTD}} & \rotatebox[origin=lb]{90}{\smash{\small Pets}} & \rotatebox[origin=lb]{90}{\smash{\small Caltech101}} & \rotatebox[origin=lb]{90}{\smash{\small Flowers}}  & \rotatebox[origin=lb]{90}{\smash{\small SUN397}} & \rotatebox[origin=lb]{90}{\smash{\small STL10}} & \rotatebox[origin=lb]{90}{\smash{\small EuroSAT}} & \rotatebox[origin=lb]{90}{\smash{\small ImageNet}} & \rotatebox[origin=lb]{90}{\smash{\small Average}} \\
\midrule
\multicolumn{17}{c}{Model Architecture: ViT-B/32} \\
\midrule

\color{dt} CLIP$^{\dag}$& \color{dt} WIT400M& \color{dt} 84.4 & \color{dt} 91.3 & \color{dt} 65.1 & \color{dt} 37.8 & \color{dt} 59.4 & \color{dt} 21.2 & \color{dt} 44.5 & \color{dt} 87.0 & \color{dt} 87.9  & \color{dt} 66.7 & \color{dt} 63.2 & \color{dt} 97.2 & \color{dt} 49.4 & \color{dt} 63.2 & \color{dt} 65.6  \\

\color{dt} CLIP$^{\ddag}$& \color{dt} WIT400M& \color{dt} 82.9 & \color{dt} 88.7 & \color{dt} 63.7 & \color{dt} 35.4 & \color{dt} 57.3 & \color{dt} 18.9 & \color{dt} 43.3 & \color{dt} 84.0 & \color{dt} 89.3  & \color{dt} 66.0 & \color{dt} 61.5 & \color{dt} 96.6 & \color{dt} 43.9 & \color{dt} 61.9 & \color{dt} 63.8  \\

OPENCLIP$^{\ddag}$&LAION400M& \bf80.5 & 90.6 & 70.7 & 42.6 & 78.1 & 15.9 & 51.3 & 85.8 & \bf91.2 & 66.0 & \bf66.7 & \bf95.2 & \bf49.7 & 62.1 & 67.6  \\
OPENCLIP$^{\star}$&LAION225M& 79.6 & 92.5 & 74.5 & 43.8 & 79.9 & 15.7 & 50.9 & 84.4 & 90.2 & 65.5 & 65.2 & 94.0 & 44.7 & 62.2 & 67.4  \\
CLIP-CID &LAION225M& 79.7 & \bf92.6 & \bf75.2 & \bf47.7 & \bf83.7 & \bf25.6 & \bf52.2 & \bf88.4 & 90.8 & \bf69.1 & 66.0 & 93.2 & 45.3 & \bf62.7 & \bf69.4 \\

\midrule
\multicolumn{17}{c}{Model Architecture: ViT-B/16}\\
\midrule
\color{dt} CLIP$^{\dag}$& \color{dt} WIT400M& \color{dt} 89.2 & \color{dt} 91.6 & \color{dt} 68.7 & \color{dt} 39.1 & \color{dt} 65.6 & \color{dt} 27.1 & \color{dt} 46.0 & \color{dt} 88.9 & \color{dt} 89.3 & \color{dt} 70.4 & \color{dt} 65.2 & \color{dt} 98.2 & \color{dt} 54.1 & \color{dt} 68.6 & \color{dt}  68.7 \\
\color{dt} CLIP$^{\ddag}$& \color{dt} WIT400M& \color{dt}87.8 & \color{dt} 89.6 & \color{dt} 66.4 & \color{dt} 40.9 & \color{dt} 63.5 & \color{dt} 23.1 & \color{dt} 44.8 & \color{dt} 87.3 & \color{dt} 90.4 & \color{dt} 67.6 & \color{dt} 63.0 & \color{dt} 98.0 & \color{dt} 52.9 & \color{dt} 67.2 & \color{dt} 67.3 \\
OPENCLIP$^{\ddag}$&LAION400M& 85.7 & 91.8 & 71.1 & 46.4 & 82.7 & 16.5 & 50.2 & \bf88.6 & 91.9 & 66.0 & \bf68.6 & \bf96.8 & 51.1 & 65.6 & 69.5 \\
OPENCLIP$^{\star}$&LAION225M& \bf85.8 &93.3 & 75.7 & 50.4 & 83.4 & 16.3 & 51.0 & 87.8 & \bf92.0 & 67.0 & 66.0 & 96.5 &  50.8 & 65.6 & 70.1 \\
CLIP-CID&LAION225M& 84.4& \bf93.9 & \bf76.9 & \bf51.5 & \bf85.2 & \bf25.5 & \bf51.7 & \bf88.6 & \bf92.0 & \bf69.0 & 66.6 & 95.1 & \bf51.3 & \bf65.8 & \bf71.3\\
\bottomrule
\end{tabular}
}
\caption{Zero-shot classification comparison. We present zero-shot performance on 14 common downstream datasets. $^{\dag}$: Results reported in CLIP paper. $^{\ddag}$: Results we reproduced. $^{\star}$: Results of the OPENCLIP model trained on LAION225M.
}
\vspace{-3mm}
\label{table:zeroshot_classification}
\end{center}
\end{table*}

\paragraph{Cluster Discrimination Distillation.} After clustering, we can inherit the original mechanism in the vanilla KD~\cite{hinton2015distilling} to implement instance-cluster alignment. As illustrated in Fig.~\ref{fig:architecture}, we consider a set of training images $I=\{x_1, x_2,...,x_n\}$, comprising $n$ images. Initially, we employ the student image encoder and the teacher image encoder to get normalized student image embeddings $E_i^s=\{e_{1}^s, e_{2}^s,..., e_{n}^s\}$ and normalized teacher image embeddings $E_i^t=\{e_{1}^t, e_{2}^t,..., e_{n}^t\}$ (the teacher image embeddings are extracted offline). These normalized image embeddings $e_{i}^s \in \mathbb{R}^{d}$ and $e_{i}^t \in \mathbb{R}^{d}$ are passed through a fully connected layer that is initialized using the cluster centers. It is important to note that without this initialization will lead to the collapse of model training. Subsequently, the images are partitioned into $k$ classes, represented by prototypes $W=\{w_i\}_{i=1}^k$. With pseudo labels and cluster centers obtained from the above clustering step, the logit distillation loss can be implemented by optimizing a standard softmax classification loss as:
\begin{equation}
\small
\begin{aligned}
\mathcal{L}_\mathrm{l} = - \sum_{i=1}^n  \log \frac{\exp(w_i^\top e_i^s)}{\sum_{j=1}^{k} \exp(w_j^\top e_i^s)}.
\end{aligned}
\end{equation}
Furthermore, we implement distribution alignment by minimizing the Kullback-Leibler (KL) divergence between the prediction probability from the teacher and the student:
\begin{equation}
\small
\begin{aligned}
&\mathcal{L}_\mathrm{d} = \sum_{i=1}^n \mathrm{KL} \left(\frac{\exp(w_i^\top e_i^s / \tau)}{\sum_{j=1}^{k} \exp(w_j^\top e_i^s/ \tau)} \bigg|\bigg| \frac{\exp(w_i^\top e_i^t / \tau)}{\sum_{j=1}^{k} \exp(w_j^\top e_i^t/ \tau)}\right),
\end{aligned}
\end{equation}
where $\tau$ is the temperature hyper-parameter used to soften distribution representation. Finally, the cluster-level distillation loss $\mathcal{L}_\mathrm{cluster}$ is defined as:
\begin{equation}
\small
\begin{aligned}
\mathcal{L}_\mathrm{cluster} = \alpha \mathcal{L}_\mathrm{l} + (1-\alpha) \mathcal{L}_\mathrm{d},
\end{aligned}
\end{equation}
where $\alpha$ is a loss weight used to balance the influence of different losses.

\subsection{Instance-level Distillation}

The cluster-level distillation primarily impacts the student image encoder, facilitating the model in capturing comprehensive semantic information from the training data. However, it may unintentionally neglect the subtle nuances of fine-grained semantic details and impose limitations on image-text alignment. To address this issue, we introduce the instance-level distillation loss. Given an image-text pair, we first get the offline extracted teacher image embedding $e_{i}^t$ and teacher text embedding $c_{i}^t$. After that, we use $e_{i}^t$ and $c_{i}^t$ to supervise the student text embedding $c_{i}^s$ and student image embedding $e_{i}^s$ respectively. We employ the bi-directional contrastive loss~\cite{sohn2016improved,srivastava2012multimodal} to align the teacher embedding and the student embedding, which is defined as:
\begin{equation}
\small
\begin{aligned}
\mathcal{L}_\mathrm{contrast}(e, c) & =\frac{1}{2}\left(\mathcal{L}_{e_{i} \rightarrow c_{i}}+\mathcal{L}_{c_{i} \rightarrow e_{i}}\right), \text { where } \\
\mathcal{L}_{e_{i} \rightarrow c_{i}} & =- \sum_{i=1}^n\log \frac{\exp({e_i^{\top} c_i / \tau})}{\sum_{j=1}^{n} \exp({e_i^{\top} c_j / \tau})}, \\
\mathcal{L}_{c_{i} \rightarrow e_{i}} & =- \sum_{i=1}^n \log \frac{\exp({e_i^{\top} c_i / \tau})}{\sum_{j=1}^{n} \exp({e_j^{\top} c_i / \tau})} .
\end{aligned}
\end{equation}

Then the instance-level distillation loss $\mathcal{L}_\mathrm{instance}$ is defined as:

\begin{equation}
\small
\begin{aligned}
\mathcal{L}_\mathrm{instance} = \gamma \mathcal{L}_\mathrm{contrast}(e_i^s, c_i^t) + (1-\gamma) \mathcal{L}_\mathrm{contrast}(c_i^s, e_i^t),
\end{aligned}
\end{equation}
where $\gamma$ is a loss weight. Finally, the overall loss function is defined as:
\begin{equation}
\small
\begin{aligned}
\mathcal{L}_\mathrm{overall} = \mathcal{L}_\mathrm{base} + \mathcal{L}_\mathrm{cluster} + \mathcal{L}_\mathrm{instance},
\end{aligned}
\end{equation}
where $\mathcal{L}_{\mathrm{base}} = \mathcal{L}_{\mathrm{contrast}}(e_i^s, c_i^s)$ represents the standard CLIP loss.

\section{Experiments and Results}
\subsection{Experimental Settings}

\noindent{\bf Implementation Details.} In this paper, we utilize the OPENCLIP bigG/14 as the teacher model. The student model adopts the same architecture as CLIP~\cite{radford2021learning}. We employ AdamW~\cite{loshchilov2017decoupled} as the optimizer, initializing it with a learning rate of $1e-3$ and a weight decay of $0.2$. To prevent collapsing, the initial learning rate of the fully connected layer is set to $1e-6$. Based on empirical observations, we set the loss weight $\alpha$ and $\gamma$ to 0.999 and 0.5 respectively. We set $\beta_{1}$ to $0.9$ and $\beta_{2}$ to $0.98$ for improved training stability. The input image size is $224 \times 224$, and the input text sequence length is truncated or padded to $77$. The temperature parameter $\tau$ is initialized to $0.07$. We conduct distillation training on ViT-B/32 and ViT-B/16 models for 32 epochs, using a batch size of $32,768$ and $24,576$ on $64$ NVIDIA H800 GPUs.

\noindent{\bf Downstream Datasets.} To prove the effectiveness of our method, we present linear probe and zero-shot classification performance across 14 datasets, including Food101~\cite{bossard2014food}, CIFAR10 \& CIFAR100~\cite{krizhevsky2009learning}, Birdsnap~\cite{berg2014birdsnap}, Stanford Cars~\cite{KrauseStarkDengFei-Fei_3DRR2013}, Aircraft~\cite{maji2013fine}, DTD~\cite{cimpoi2014describing}, Oxford Pets~\cite{parkhi2012cats}, Caltech101~\cite{fei2004learning}, Flowers102~\cite{nilsback2008automated}, SUN397~\cite{xiao2010sun}, STL10~\cite{coates2011analysis}, EuroSAT~\cite{helber2019eurosat}, and ImageNet~\cite{deng2009imagenet}. Additionally, to evaluate the robustness of our model, we conduct zero-shot robustness comparison on ImageNet-V2~\cite{recht2019imagenet}, ImageNet-A~\cite{recht2019imagenet}, ImageNet-R~\cite{hendrycks2021many}, ObjectNet~\cite{barbu2019objectnet}, and ImageNet-Sketch~\cite{wang2019learning}.

\subsection{Experimental Results}

\begin{table*}[t]
\begin{center}

\resizebox{\textwidth}{!}{
\begin{tabular}{l@{\hspace{1.1em}}c@{\hspace{0.5em}}c@{\hspace{0.6em}}c@{\hspace{0.6em}}c@{\hspace{0.6em}}c@{\hspace{0.6em}}c@{\hspace{0.6em}}c@{\hspace{0.6em}}c@{\hspace{0.6em}}c@{\hspace{0.6em}}c@{\hspace{0.6em}}c@{\hspace{0.6em}}c@{\hspace{0.6em}}c@{\hspace{0.6em}}c@{\hspace{0.6em}}c@{\hspace{0.6em}}c@{\hspace{0.6em}}}
\toprule
Model& Dataset & \rotatebox[origin=lb]{90}{\smash{\small Food101}} & \rotatebox[origin=lb]{90}{\smash{\small CIFAR10}} & \rotatebox[origin=lb]{90}{\smash{\small CIFAR100}} & \rotatebox[origin=lb]{90}{\smash{\small Birdsnap}} & \rotatebox[origin=lb]{90}{\smash{\small  Cars}} & \rotatebox[origin=lb]{90}{\smash{\small Aircraft}} & \rotatebox[origin=lb]{90}{\smash{\small DTD}} & \rotatebox[origin=lb]{90}{\smash{\small Pets}} & \rotatebox[origin=lb]{90}{\smash{\small Caltech101}} & \rotatebox[origin=lb]{90}{\smash{\small Flowers}}  & \rotatebox[origin=lb]{90}{\smash{\small SUN397}} & \rotatebox[origin=lb]{90}{\smash{\small STL10}} & \rotatebox[origin=lb]{90}{\smash{\small EuroSAT}} & \rotatebox[origin=lb]{90}{\smash{\small ImageNet}} & \rotatebox[origin=lb]{90}{\smash{\small Average}} \\
\midrule
\multicolumn{17}{c}{Model Architecture: ViT-B/32} \\
\midrule

\color{dt} CLIP$^{\dag}$& \color{dt} WIT400M& \color{dt} 88.8 & \color{dt} 95.1 & \color{dt} 80.5 &\color{dt} 58.5 & \color{dt} 81.8 & \color{dt} 52.0 & \color{dt} 76.5 & \color{dt} 90.0 & \color{dt} 93.0  & \color{dt} 96.9 &  \color{dt} 76.6 & \color{dt} 98.3 &\color{dt} 97.0 & \color{dt} 76.1 & \color{dt} 82.9  \\
\color{dt}CLIP$^{\ddag}$& \color{dt} WIT400M&\color{dt} 88.6 & \color{dt} 95.0 & \color{dt} 80.2 &  \color{dt} 61.8 & \color{dt} 81.3 &  \color{dt} 50.6 & \color{dt} 76.4 & \color{dt} 89.3 & \color{dt} 92.7 & \color{dt} 94.6 & \color{dt} 76.9 & \color{dt} 98.2 & \color{dt} 95.0 & \color{dt} 75.0 & \color{dt} 82.5 \\
OPENCLIP$^{\ddag}$&LAION400M& \bf 87.0 & 95.9 & 83.0 & 64.1 & 89.5 & 54.3 & 79.8 & 88.8 & 93.4 & 95.8 & \bf 77.0 & 97.8 & 95.9 & 73.8 & 84.0  \\
OPENCLIP$^{\star}$&LAION225M& 86.2 & 96.7 & 85.0 & 62.8 & 90.0 & 55.4 & 80.3 & 88.4 & 93.1 & 95.7 & 76.5 & 97.7 & 95.8 & 74.0 & 84.1  \\
CLIP-CID&LAION225M& 86.6 & \bf97.1 & \bf86.1 & \bf70.0 & \bf 93.0 & \bf 65.1 & \bf 80.6 & \bf91.6 & \bf 94.6 & \bf 97.2 & 76.5 & \bf 98.1 & \bf96.3 & \bf 75.1 & \bf 86.3 \\

\midrule
\multicolumn{17}{c}{Model Architecture: ViT-B/16}\\
\midrule
\color{dt} CLIP$^{\dag}$& \color{dt} WIT400M& \color{dt} 92.8 & \color{dt} 96.2 & \color{dt} 83.1 & \color{dt} 67.8 & \color{dt} 86.7 & \color{dt} 59.5 & \color{dt} 79.2 & \color{dt} 93.1 & \color{dt} 94.7 & \color{dt} 98.1 & \color{dt} 78.4 & \color{dt} 99.0 & \color{dt} 97.1 & \color{dt} 80.2  & \color{dt} 86.1 \\
\color{dt}CLIP$^{\ddag}$& \color{dt} WIT400M& \color{dt}92.7 & \color{dt} 96.0 & \color{dt} 82.5 & \color{dt} 72.4 & \color{dt} 86.4 & \color{dt} 59.7 & \color{dt} 78.9 & \color{dt} 93.1 & \color{dt} 94.0 & \color{dt} 96.5 & \color{dt} 78.6 & \color{dt} 99.1 & \color{dt} 95.3 & \color{dt} 79.3 & \color{dt} 86.0 \\
OPENCLIP$^{\ddag}$&LAION400M& \bf 90.8 & 96.4 & 84.0 & 71.7 & 92.0 & 59.5 & 81.5 & 91.7 & 94.7 & 97.3 & \bf 79.4 & \bf 98.7 & 96.1 & 77.7 & 86.5\\
OPENCLIP$^{\star}$&LAION225M& 90.1 & 97.3 & 86.4 & 73.2 & 92.8 & 59.0 & 81.4 & 91.7 & 94.6 & 96.9 & 78.5 & 98.5 &  \bf 96.8 & 77.6 & 86.8 \\
CLIP-CID&LAION225M& 90.5 & \bf97.4 & \bf87.2 & \bf77.2 & \bf 93.8 & \bf 70.9 & \bf81.8 & \bf92.9 & \bf 95.0 & \bf 98.4 & 78.3 & 98.6 & \bf96.8 & \bf78.0 & \bf88.3\\
\bottomrule
\end{tabular}
}
\caption{Linear probe comparison. We present linear probe performance on 14 common downstream datasets. $^{\dag}$: Results reported in CLIP paper. $^{\ddag}$: Results we reproduced. $^{\star}$: Results of the OPENCLIP model trained on LAION225M.
}
\label{table:linear}
\end{center}
\vspace{-3mm}
\end{table*}

\noindent{\bf Zero-shot Classification.} We present our performance on 14 zero-shot classification datasets. The prompt templates and class names are consistent with CLIP~\cite{radford2021learning}. As shown in Tab.~\ref{table:zeroshot_classification}, the OPENCLIP trained on LAION225M achieves comparable performance with trained on LAION400M. This is primarily attributed to the removal of semantically redundant images, which enhances both the semantic balance and diversity of the pre-training dataset. Furthermore, through the integration of cluster-instance discrimination distillation, CLIP-CID ViT-B/32 and CLIP-CID ViT-B/16 achieve an average performance of $69.4\%$ and $71.3\%$ respectively across the 14 datasets, surpassing OPENCLIP trained on LAION400M by $1.8\%$ and $1.8\%$. This performance improvement demonstrates the effectiveness of the cluster-instance discrimination distillation in enhancing student representations.

\begin{table}[t!]
\begin{center}

\resizebox{\linewidth}{!}{
\begin{tabular}{l@{\hspace{1.1em}}c@{\hspace{0.5em}}c@{\hspace{0.6em}}c@{\hspace{0.6em}}c@{\hspace{0.6em}}c@{\hspace{0.6em}}c@{\hspace{0.6em}}c@{\hspace{0.6em}}}
\toprule
Model& Dataset & \small IN-V2 & \small IN-A & \small IN-R & \small Object & \small IN-S & \small Average\\
\midrule
\multicolumn{8}{c}{Model Architecture: ViT-B/32} \\
\midrule
\color{dt} CLIP$^{\ddag}$& \color{dt} WIT400M& \color{dt} 54.9 & \color{dt} 31.9 & \color{dt} 67.1 & \color{dt} 52.6 &\color{dt} 38.8 & \color{dt} 49.1  \\
OPENCLIP$^{\ddag}$&LAION400M& 54.6 & \bf 22.2 & 72.3 & 52.7 & 46.5 & 49.7  \\
OPENCLIP$^{\star}$&LAION225M& 54.0 & 21.2  & 71.4 & 51.3 & 46.5 & 48.9   \\
CLIP-CID &LAION225M& \bf 54.7 &  21.9 & \bf 72.5 &\bf 52.9 &\bf 47.1 & \bf 49.8  \\
\midrule
\multicolumn{8}{c}{Model Architecture: ViT-B/16} \\
\midrule
\color{dt} CLIP$^{\ddag}$& \color{dt} WIT400M & \color{dt} 61.1 &  \color{dt} 49.6 &  \color{dt} 76.0 & \color{dt} 59.0 & \color{dt} 45.1 & \color{dt} 58.1 \\
OPENCLIP$^{\ddag}$ & LAION400M & 59.1 & \bf 33.4 & 76.8 & 57.0 & 49.8 & 55.2  \\
OPENCLIP$^{\star}$ & LAION225M & 58.2 & 31.4 & 76.5 & 55.7 & 50.3 & 54.4  \\
CLIP-CID & LAION225M & \bf59.3 & 32.9 & \bf77.1 & \bf 57.5 & \bf 50.5 & \bf 55.5  \\
\bottomrule
\end{tabular}
}
\caption{Zero-shot robustness comparison. $^{\ddag}$: Results we reproduced. $^{\star}$: Results of the OPENCLIP model trained on LAION225M.}
\label{table:robustness}
\end{center}
\vspace{-5mm}
\end{table}

\noindent{\bf Linear Probe.} Following the same evaluation setting as CLIP~\cite{radford2021learning}, we freeze our model and only train a logistic regression classifier. In Tab.~\ref{table:linear}, we present linear probe performance on 14 downstream datasets. Similarity with zero-shot classification, we observe that after removing 43.7\% of the training data, the OPENCLIP trained on the filtered LAION225M achieves similar performance with trained on the entire LAION400M. By employing our proposed cluster-instance discrimination distillation method, CLIP-CID demonstrates an average performance improvement of $2.3\%$ and $1.8\%$ across 14 datasets. Notably, our method exhibits superior performance on CIFAR10\&CIFAR100, Oxford Pets, Birdsnap, Stanford Car, and Aircraft, further substantiating the effectiveness of our approach in significantly enhancing the representation power for instance discrimination.

\begin{figure}[t!]
\centering
  \begin{subfigure}{0.22\textwidth}
    \includegraphics[width=\textwidth]{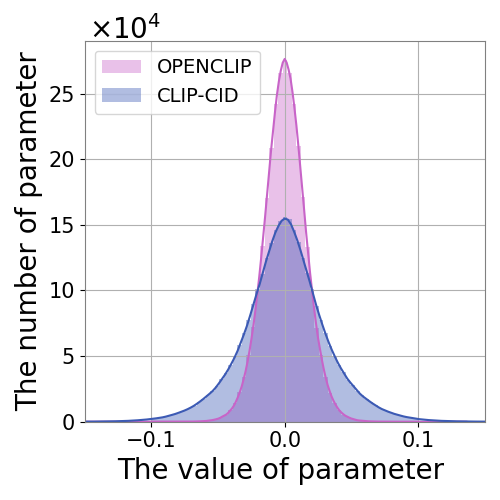}
    \caption{\small Middle Transformer Layer}
  \end{subfigure}
  \begin{subfigure}{0.22\textwidth}
    \includegraphics[width=\textwidth]{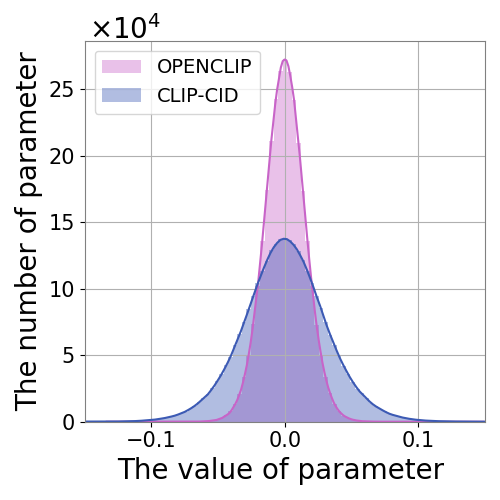}
    \caption{\small Last Transformer Layer}
  \end{subfigure}
\caption{Weight distribution of the last fully connected layer in the middle and last transformer layers.}
\vspace{-3mm}
\label{fig:fc_weight}
\end{figure}

\noindent{\bf Zero-shot Robustness Evaluation.} In Tab.~\ref{table:robustness}, we present a robustness evaluation across different model sizes. We observe that excluding 43.7\% of the training data leads to a marginal decrease in robustness. Subsequent integration of the cluster-instance discrimination distillation mechanism enables our model to acquire knowledge from the teacher model efficiently and effectively. Consequently, our model consistently exhibits superior robustness compared to OPENCLIP. In Fig.~\ref{fig:fc_weight}, we visually depict the weight distribution of the last fully connected layer in both the middle and last transformer layers. After distillation, we observe an expanded range of values within the weight distribution of our model, accompanied by a reduction in the number of elements proximal to zero. This phenomenon reflects the improvement of capacity since the weights can assume a more diverse array of potential values or states~\cite{shen2020meal}.

\subsection{Ablation Study}

\noindent{\bf Ablation on Threshold $\bm{\beta}$.} To explore the optimal image filtering ratio of the LAION400M, we perform an ablation study on threshold $\beta$. The value of $\beta$ is associated with the number of sets, which directly influences the proportion of removed images. We train standard OPENCLIP ViT-B/32 on the filtered dataset with various values of the threshold $\beta$. As shown in Tab.~\ref{table:ablation_threshold}, setting $\beta$ to 0.07 results in the removal of 43.7\% of image-text pairs, and we observe the optimal performance. However, increasing $\beta$ to 0.08 raised the filtration rate to 47.5\%, leading to a significant performance decline. 

\begin{table}[t!]
    \centering
    \resizebox{\linewidth}{!}{
    \begin{tabular}{lcc|cc}
    \toprule
    $\beta$ & Dataset & Filtration Ratio & Linear probe & Zero-shot \\
    \midrule
    0.06 & LAION237M & 40.8\%& 83.4 & 66.2 \\
    0.07 & LAION225M & 43.7\%&\bf 84.1 & \bf 67.4 \\
    0.08 & LAION210M & 47.5\%&83.7 & 67.1 \\
    \bottomrule
    \end{tabular}
    }
    \caption{Ablation on different threshold $\beta$.}
    \vspace{-3mm}
    \label{table:ablation_threshold}
\end{table}

\noindent{\bf Ablation on Distillation Loss.} We perform ablation experiments on distillation losses to validate our proposed cluster-instance distillation mechanism. As shown in Tab.~\ref{table:ablation_loss}, the integration of cluster discrimination distillation loss improves linear probe performance from 84.1\% to 85.7\%, while the marginal gain in zero-shot performance is only 0.9\%. This is because the cluster-level distillation loss mainly affects the image encoder, enhancing its ability to learn comprehensive semantic information while imposing constraints on image alignment. After introducing the instance discrimination distillation loss, it improves cross-modal alignment and captures more fine-grained semantics, which significantly boosts the zero-shot result from 68.3\% to 69.4\%.

\begin{table}[t]
    \centering
    \resizebox{\linewidth}{!}{
    \begin{tabular}{cccc|cc}
    \toprule
    $\mathcal{L}_\mathrm{base}$& $\mathcal{L}_\mathrm{cluster}$ & $\mathcal{L}_\mathrm{instance}$ & Dataset & Linear probe  & Zero-shot\\
    \midrule
    \Checkmark & \XSolidBrush &\XSolidBrush &LAION225M& 84.1  & 67.4  \\
    \Checkmark & \Checkmark  & \XSolidBrush &LAION225M& 85.7  & 68.3  \\
    \Checkmark & \XSolidBrush  & \Checkmark &LAION225M& 85.3  & 68.8  \\
    \Checkmark & \Checkmark & \Checkmark &LAION225M& \bf86.3  & \bf69.4  \\
    \bottomrule
    \end{tabular}
    }
\caption{Ablation on different loss combinations.}
\vspace{-3mm}
\label{table:ablation_loss}
\end{table}

\noindent{\bf Ablation on Cluster Centers.} The number of cluster centers is a vital factor in managing inter-class and intra-class conflicts. As demonstrated in Tab.~\ref{table:ablation_cluster}, we present the average linear probe and zero-shot classification performance of CLIP-CID ViT-B/32. The increase in the number of cluster centers from 0.1 million to 1 million resulted in a corresponding enhancement in model performance. This enhancement can be attributed to the increased intra-cluster purity, signifying improved discrimination and representation capabilities of the model. However, due to increased inter-cluster conflicts, the performance deteriorates as the number of cluster centers rises from 1M to 5M.

\begin{table}[h!]
    \centering
    \resizebox{1\linewidth}{!}{
    \begin{tabular}{lc|cc}
        \toprule
        Cluster Centers & Dataset & Linear probe & Zero-shot \\
        \midrule
        0.1M & LAION225M & 83.8 & 67.8 \\
        0.5M & LAION225M & 84.6 & 68.3 \\
        1M & LAION225M & \bf 86.3 & \bf 69.4 \\
        2M & LAION225M & 85.8 & 69.0 \\
        5M & LAION225M & 85.2 & 68.8 \\
        \bottomrule
    \end{tabular}
   }
    \caption{Ablation on different numbers of cluster centers. }
    \label{table:ablation_cluster}
    \vspace{-3mm}
\end{table}

\noindent{\bf Ablation on Teacher Models.}
To explore the impact of the difference in parameter quantity between teacher and student models, we compare the performance of CLIP-CID ViT-B/32 distilled from different scale teacher models.
The experiment results are shown in Tab.~\ref{table:ablation_teacher}, we find a larger teacher model such as OPENCLIP ViT-bigG/14 produces better student performance for both linear probe and zero-shot classification.
\begin{table}[t!]
    \centering
    \resizebox{1\linewidth}{!}{
    \begin{tabular}{lc|cc}
    \toprule
    Teacher Model  & Dataset & Linear probe & Zero-shot \\
    \midrule
    OPENCLIP ViT-L/14 & LAION225M  & 85.5 & 68.7 \\
    OEPNCLIP ViT-bigG/14 & LAION225M  & \bf 86.3 & \bf69.4 \\
    \bottomrule
    \end{tabular}
    }
    \caption{Ablation on different scales of teacher models.}
    \label{table:ablation_teacher}
    \vspace{-3mm}
\end{table}


\begin{figure}[t!]
 \centering
  \includegraphics[width=\linewidth]{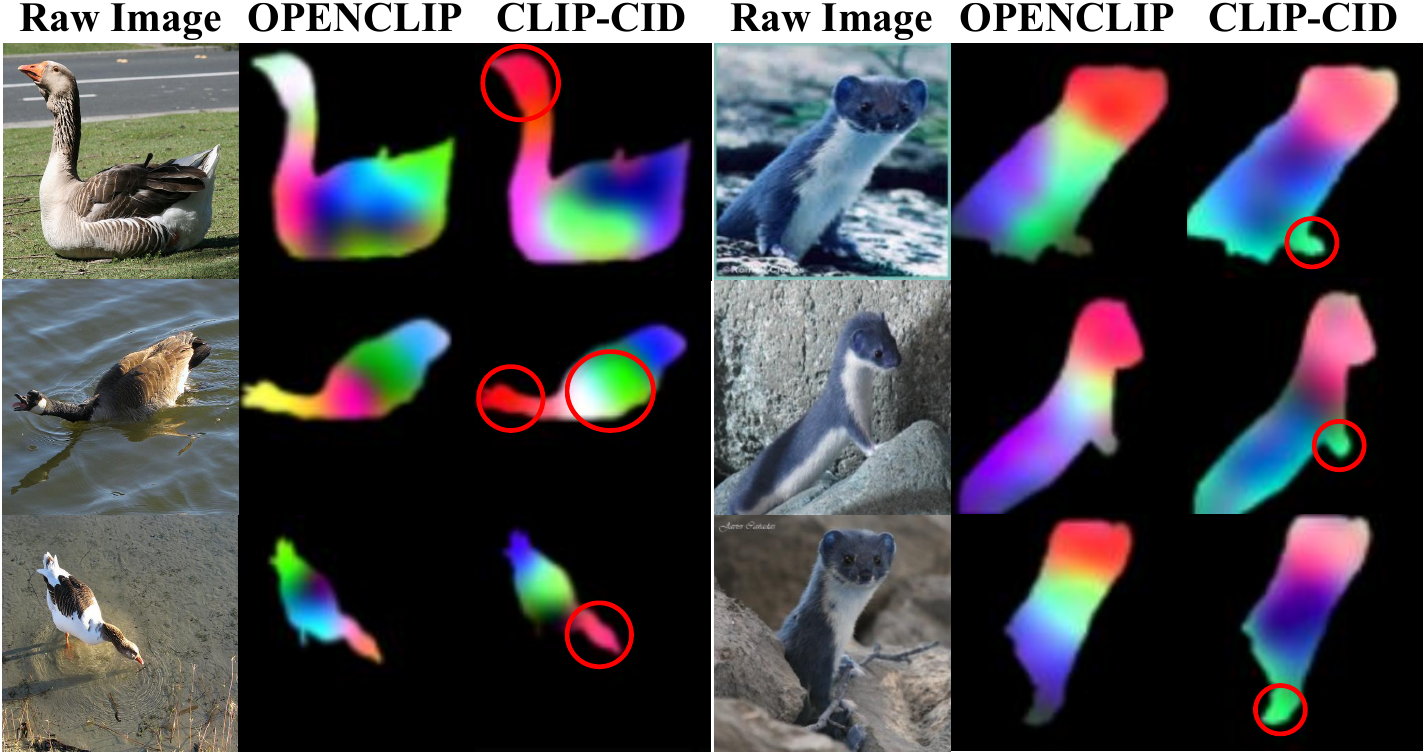}
  \caption{Visualization of PCA components. We extract three principal components from the collected patch features of each image. The principal components are then visualized using separate color channels. Similar colors within patches indicate semantic similarities. We use \scalebox{2}{$\textcolor{red}{\circ}$} to accentuate the primary distinction.}
  \label{fig:pca}
\vspace{-5mm}
\end{figure}

\noindent{\bf Visualization of PCA Components.} We present the results of Principal Component Analysis (PCA) applied to patch features extracted by OPENCLIP ViT-B/32 and our CLIP-CID ViT-B/32. We retain patches with positive values after applying a threshold to the first component, effectively separating the main object from the background. Subsequently, a second PCA is computed on the remaining patches. To visualize the results, we assign three distinct colors to the first three components of each model. In Fig.~\ref{fig:pca}, our model demonstrates superior semantic understanding by consistently preserving the consistent color representation of object parts across diverse images within the same category. For example, the visualization of our model consistently maintains the head of the goose in a consistent red color. However, the OPENCLIP displays three different colors. We also provide visualization of embeddings, clusters, and class activation maps, please refer to the supplementary material.

\subsection{Conclusion}

In this paper, we introduce CLIP-CID, a novel distillation mechanism that effectively transfers knowledge from a large vision-language foundation model to a smaller model. To mitigate transfer learning bias and enhance distillation efficiency, we propose an efficient image semantic balance method, which can filter 43.7\% of image-text pairs from the LAION400M while maintaining superior performance. After that, we employ cluster-instance discrimination to facilitate knowledge transfer from the teacher to the student model, enabling the student model to achieve a comprehensive semantic understanding of the pre-training data. Experimental results demonstrate that CLIP-CID surpasses existing methods in various downstream tasks, including linear probe and zero-shot classification.

\bibliography{aaai25}

\appendix

\section{Detail Experimental Settings}

\subsection{Experimental Settings}
In Tab.~\ref{table:hyperparam}, we present the detailed experimental settings used in the distillation process.

\begin{table}[h!]
\centering
\resizebox{0.6\linewidth}{!}{
\begin{tabular}{l|c}
\toprule Hyperparameter & Value \\
\midrule
Initial temperature & $0.07$ \\
Adam $\beta_{1}$ & $0.9$ \\
Adam $\beta_{2}$ & $0.98$ \\
Adam $\epsilon$  & $10^{-6}$ \\
Weight decay & $0.2$ \\
Batch size & 32768 \\
Learning rate & 0.001 \\
Learning rate scheduler & OneCycleLR \\
Pct start & 0.1 \\
Training epochs & 32  \\
GPU & $64 \times $H800 \\
\bottomrule
\end{tabular}
}
\caption{Hyperparameters used in the distillation process.}
\label{table:hyperparam}
\vspace{-5mm}
\end{table}

\subsection{Downstream Datasets}

We use 14 image classification datasets to prove the effectiveness of our method, including Food101~\cite{bossard2014food}, CIFAR10 \& CIFAR100~\cite{krizhevsky2009learning}, Birdsnap~\cite{berg2014birdsnap}, Stanford Cars~\cite{KrauseStarkDengFei-Fei_3DRR2013}, Aircraft~\cite{maji2013fine}, DTD~\cite{cimpoi2014describing}, Oxford Pets~\cite{parkhi2012cats}, Caltech101~\cite{fei2004learning}, Flowers102~\cite{nilsback2008automated}, SUN397~\cite{xiao2010sun}, STL10~\cite{coates2011analysis}, EuroSAT~\cite{helber2019eurosat}, and ImageNet~\cite{deng2009imagenet}. Details on each dataset and the corresponding evaluation metrics are provided in Tab.~\ref{linearprobedatasets}.

\begin{table}[h!]
\centering

\resizebox{\linewidth}{!}{
\begin{tabular}{lcccr}
\toprule
\multicolumn{1}{l}{Dataset} & \multicolumn{1}{c}{Classes} & \multicolumn{1}{c}{Train size} & \multicolumn{1}{c}{Test size} & \multicolumn{1}{c}{Evaluation metric} \\
\midrule
Food101 & 102 & 75,750 & 25,250 & accuracy \\
CIFAR10 & 10 & 50,000 & 10,000 & accuracy \\
CIFAR100 & 100 & 50,000 & 10,000 & accuracy \\
Birdsnap & 500 & 42,283 & 2194 & accuracy \\
Stanford Cars & 196 & 8,144 & 8,041 & accuracy \\
Aircraft & 100 & 6,667 & 3,333 & mean per class \\
DTD & 47 & 3,760 & 1,880 & accuracy \\
Oxford Pets & 37 & 3,680 & 3,669 & mean per class \\
Caltech101 & 102 & 3,060 & 6,085 & mean per class \\
Flowers102 & 102 & 2,040 & 6,149 & mean per class \\
SUN397 & 397 & 19,850 & 19,850 & accuracy \\
STL10 & 10 & 1,000 & 8,000 & accuracy \\
EuroSAT & 10 & 10,000 & 5,000 & accuracy \\
ImageNet & 1,000 & 1,281,167 & 50,000 & accuracy \\
\bottomrule
\end{tabular}}
\caption{List of the linear probe and zero-shot classification datasets with the data distribution and evaluation metrics.}
\label{linearprobedatasets}
\vspace{-3mm}
\end{table}

\begin{figure}[t!]
 \centering
  \includegraphics[width=\linewidth]{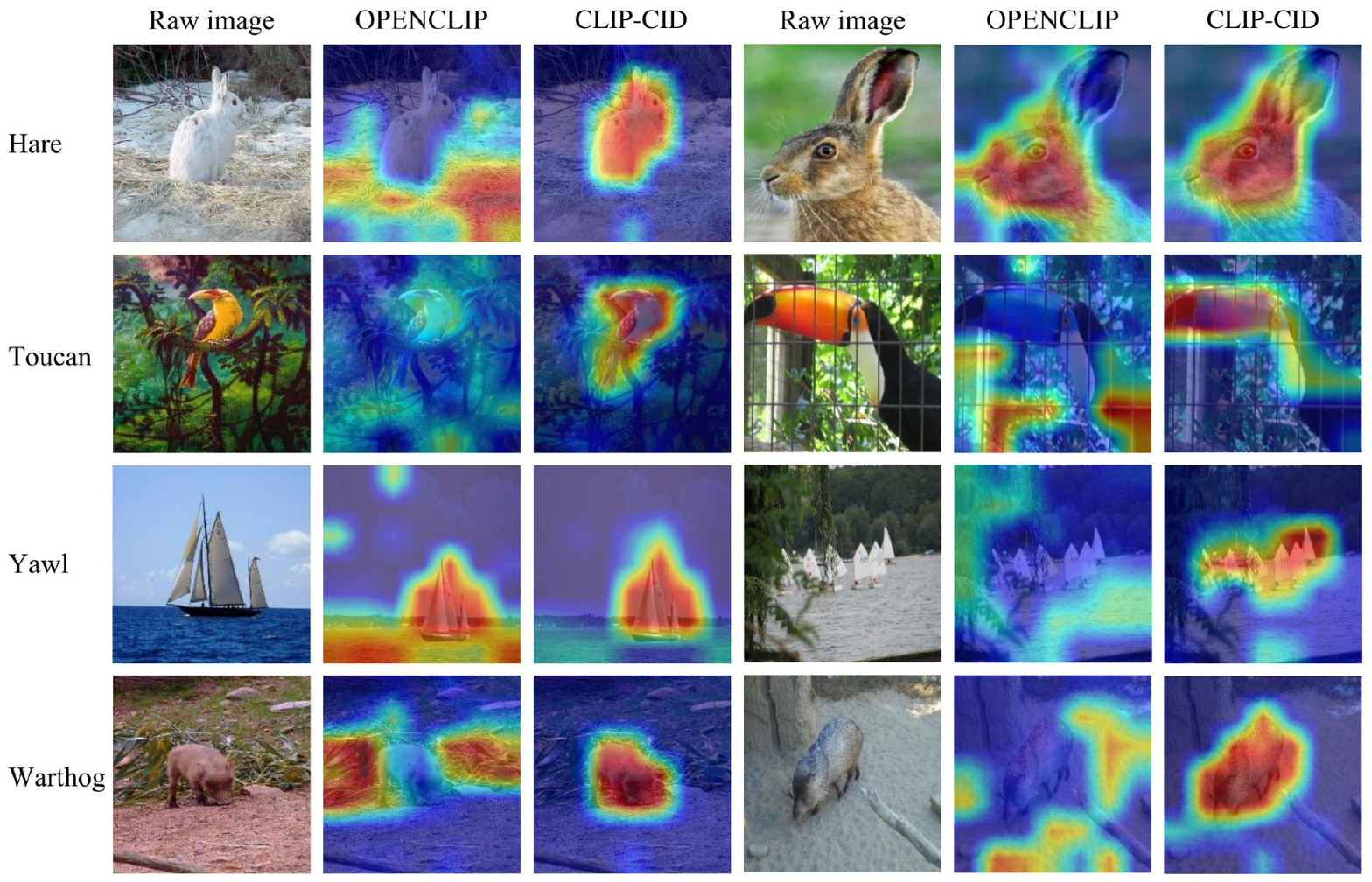}
  \vspace{-5mm}
  \caption{Class activation maps for OPENCLIP and CLIP-CID on different classes from ImageNet.}
  \label{fig:classactivate}
  \vspace{-5mm}
\end{figure}

\subsection{Model Architectures}
We follow the same architecture design as OPENCLIP. The details model paramters of CLIP-CID ViT-B/32 and ViT-B/16 are present in Tab.~\ref{tab:alip_model_Hyperparameter}.

\section{More Visualization}

\subsection{Class Activation Maps}
In Fig.~\ref{fig:classactivate}, we present class activation maps of OPENCLIP and our model for different classes from ImageNet. Beneficial from the cluster-instance discrimination distillation, our proposed CLIP-CID is superior in effectively aligning image patches and textual tokens. 

\begin{figure}[h!]
\centering
  \begin{subfigure}{0.32\linewidth}
    \includegraphics[width=1.0\linewidth]{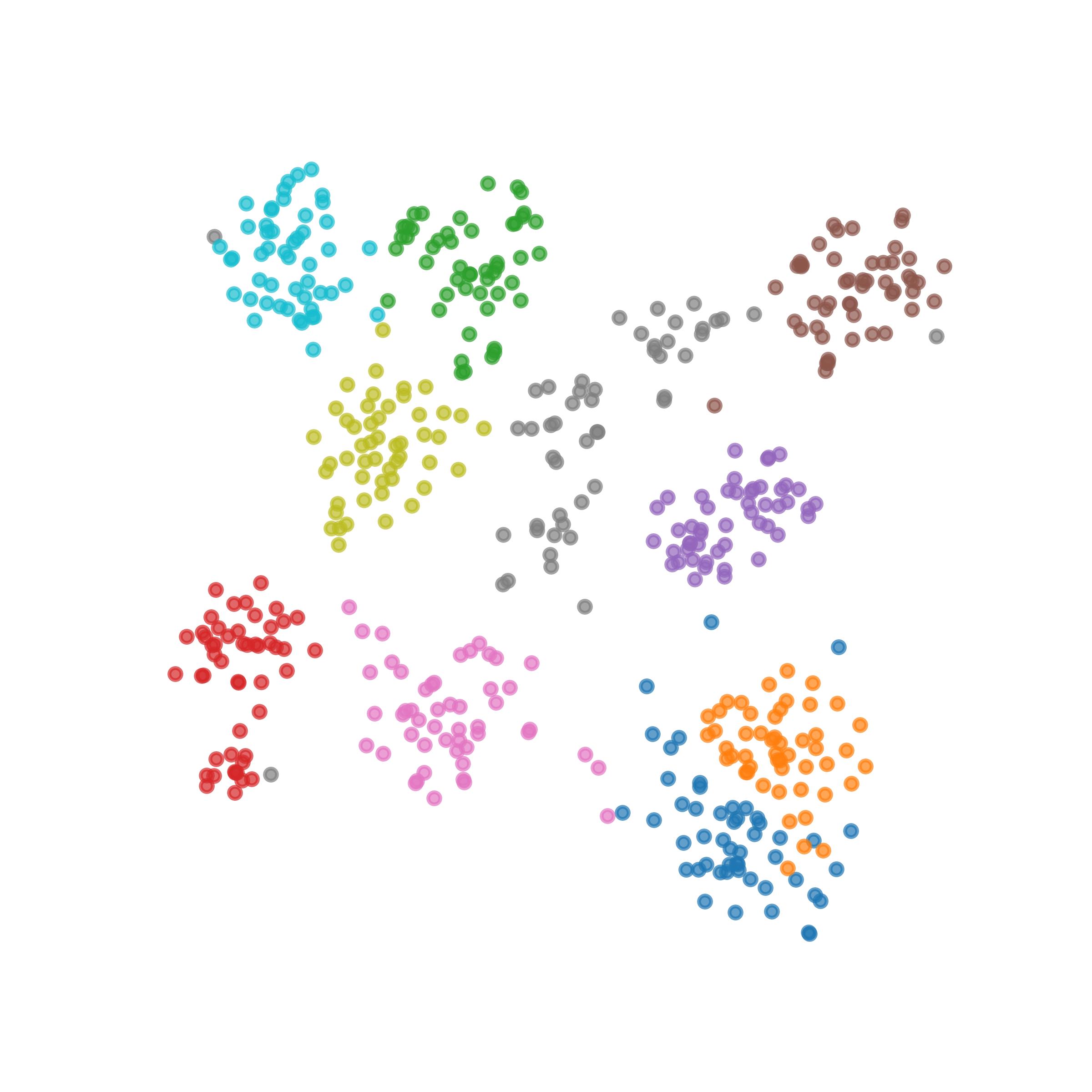}
    \caption{OPENCLIP} 
  \end{subfigure}
  \begin{subfigure}{0.32\linewidth}
    \includegraphics[width=1.0\linewidth]{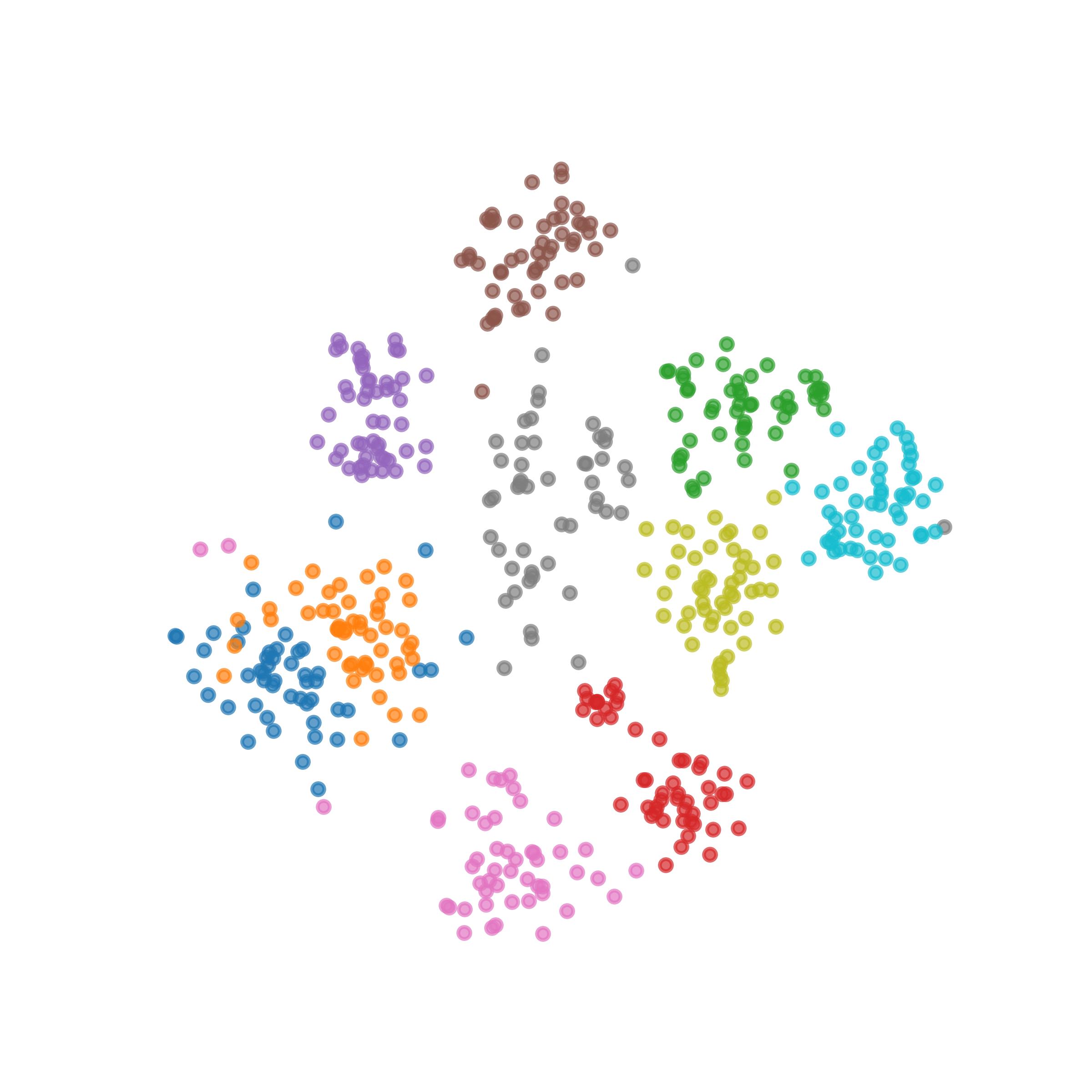}
    \caption{OPENCLIP$^{\star}$} 
  \end{subfigure}
  \begin{subfigure}{0.32\linewidth}
    \includegraphics[width=1.0\linewidth]{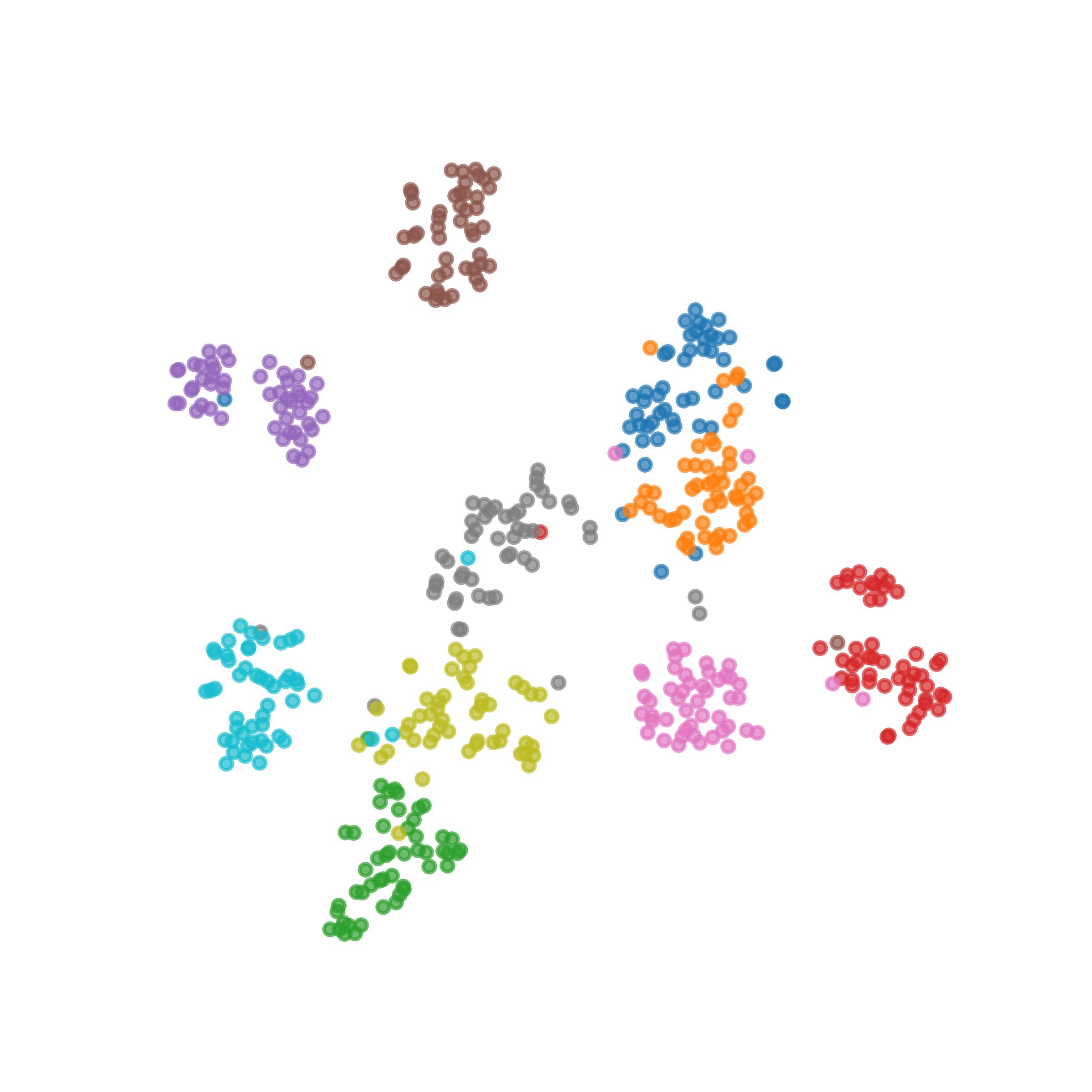}
    \caption{CLIP-CID} 
  \end{subfigure}
  \caption{Visualization of validation images from the ImageNet by t-SNE. We randomly sample 10 classes within 1000 classes. The distilled CLIP-CID learns more compact and more discriminative representations. $^{\star}$: Results of the OPENCLIP model trained on LAION225M.}
\label{fig:tSNE1}
\vspace{-5mm}
\end{figure}

\subsection{Embedding Visualiztion}
To visually illustrate the differences between our distilled model and the OPENCLIP model at the feature level, we employ t-SNE to visualize the embeddings. We randomly select 10 classes from the validation dataset of ImageNet. The visualization results are shown in Fig.~\ref{fig:tSNE1}. In contrast to training on LAION400M, OPENCLIP ViT-B/32 trained on the image semantic balanced LAION225M demonstrates similar competencies. Furthermore, the distilled CLIP-CID adeptly captures semantic details through cluster-instance discrimination, enhancing intra-cluster cohesion and inter-cluster discrimination.

\begin{table*}[h!]
    \centering
    \resizebox{0.9\linewidth}{!}{
    \begin{tabular}{l|cc cccc ccc}
        \toprule 
        \multirow{2}{*}{ Model } & Embedding & Input & \multicolumn{4}{c}{ Image Encoder } & \multicolumn{3}{c}{ Text Encoder } \\
          & dimension & resolution & layers & width & heads & patches & layers & width & heads \\
        \midrule 
        ViT-B/32  &  512 & $224\times 224$ & 12 & 768 & 12 & 32 & 12 & 512 & 8 \\
        ViT-B/16 &  512 & $224\times 224$ & 12 & 768 & 12 & 16 & 12 & 512 & 8 \\
        \bottomrule
    \end{tabular}
    }
\caption{The detailed architecture parameters for our proposed CLIP-CID.}
\label{tab:alip_model_Hyperparameter}
\vspace{-3mm}
\end{table*}

\begin{figure*}[h!]
\centering
  \begin{subfigure}{0.31\textwidth}
    \includegraphics[width=0.98\textwidth]{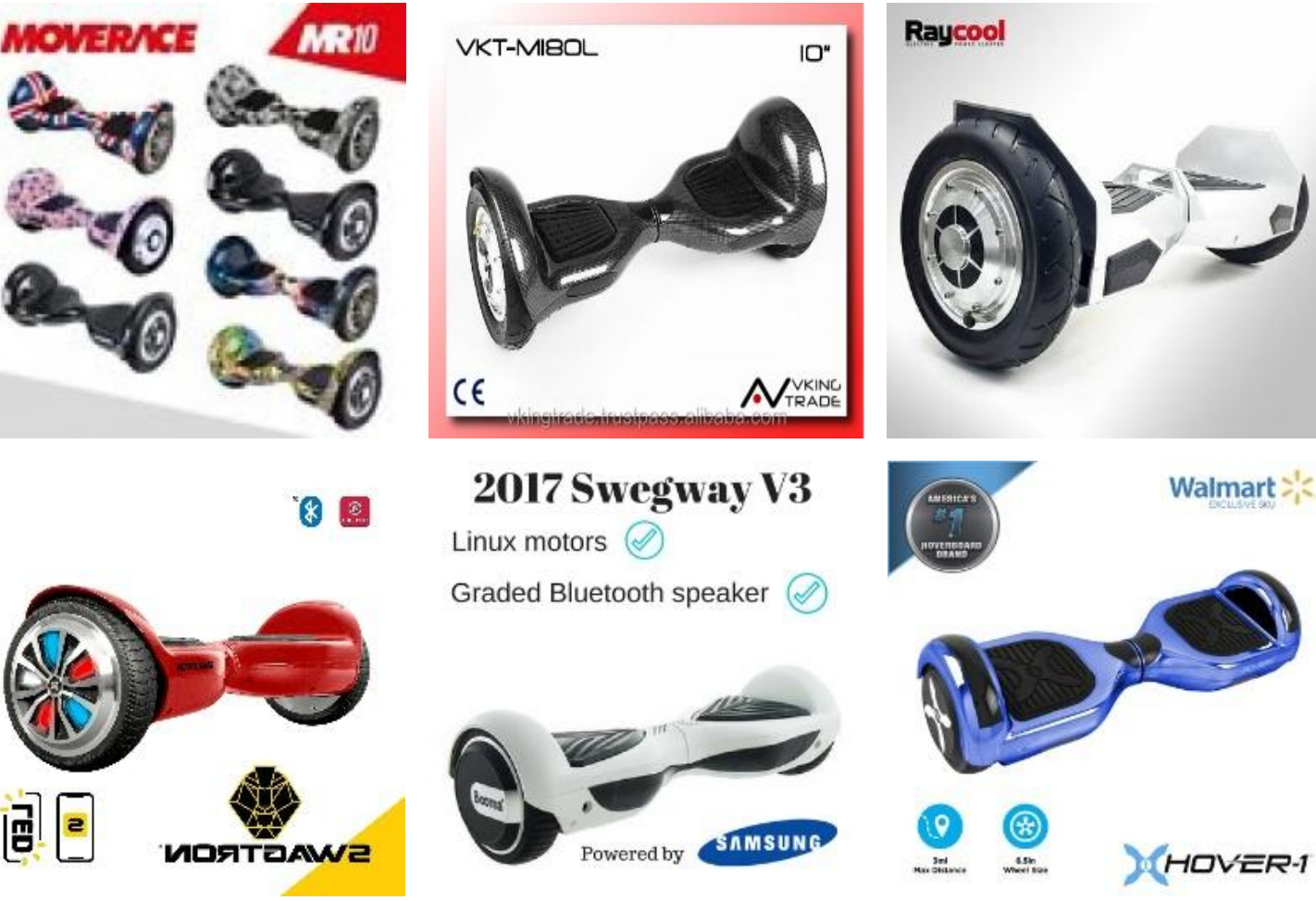}
  \end{subfigure}
  \begin{subfigure}{0.31\textwidth}
    \includegraphics[width=0.98\textwidth]{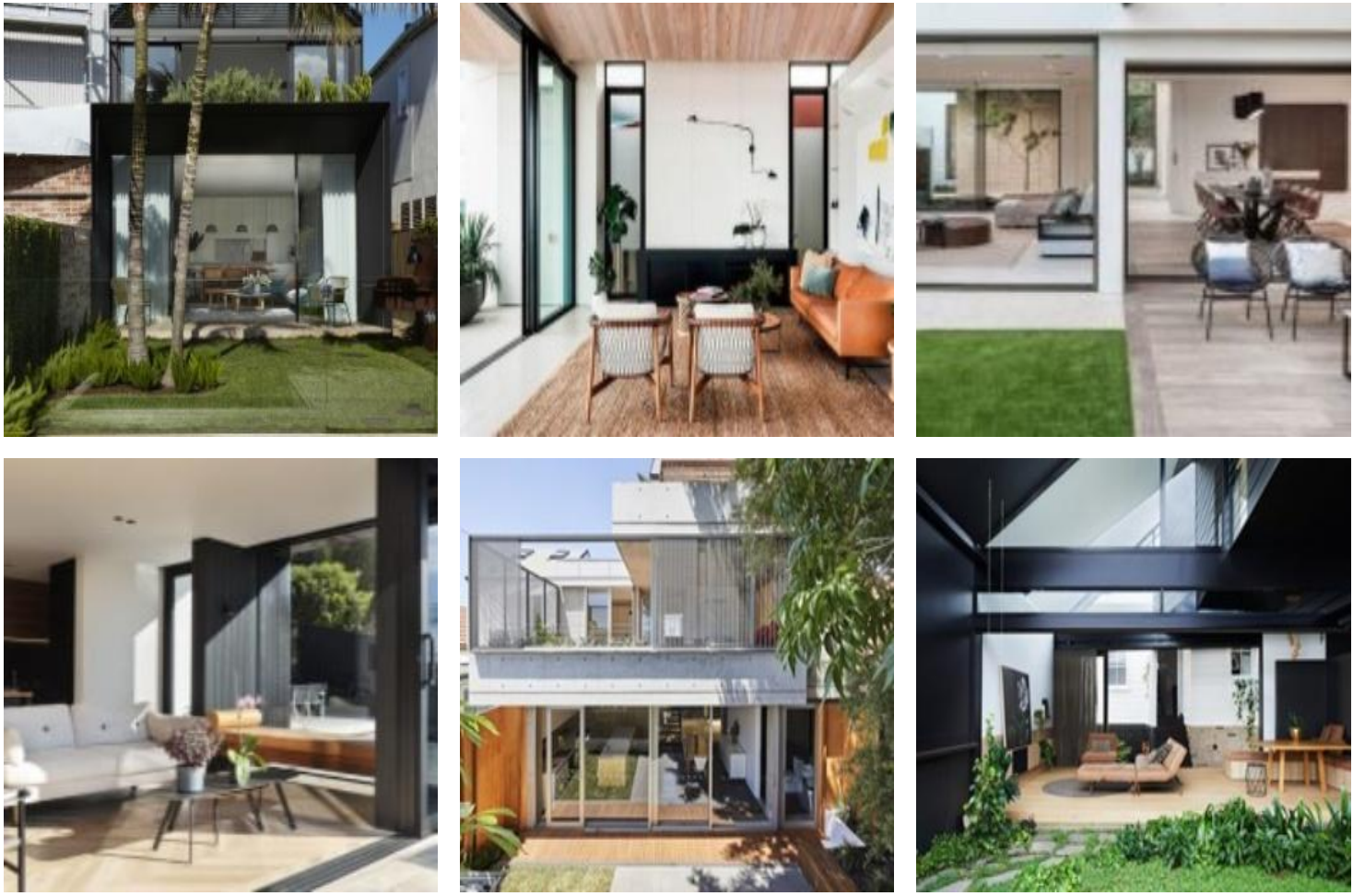}
  \end{subfigure}
  \begin{subfigure}{0.31\textwidth}
    \includegraphics[width=0.98\textwidth]{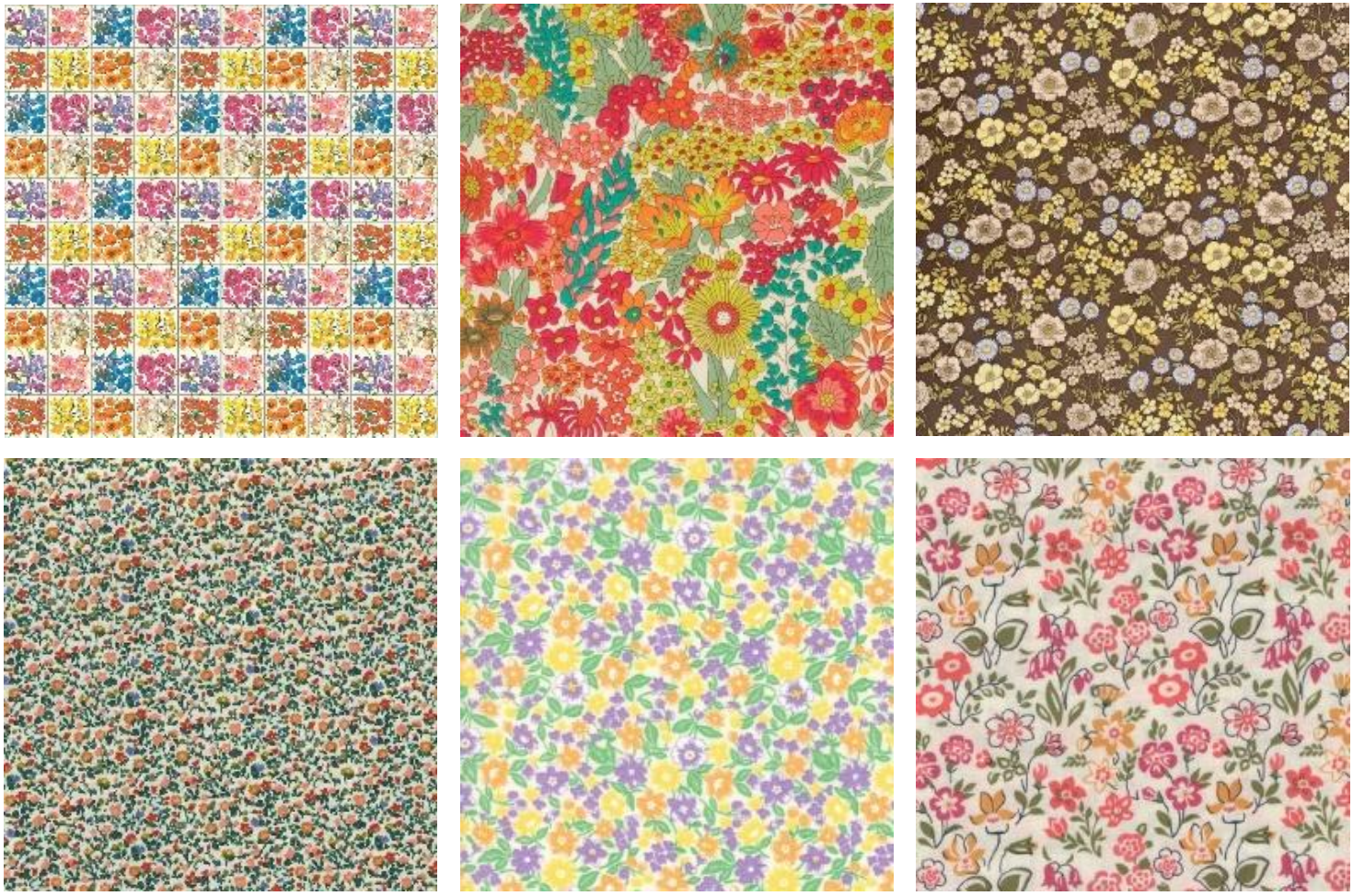}
  \end{subfigure}
  
  \begin{subfigure}{0.31\textwidth}
    \includegraphics[width=0.98\textwidth]{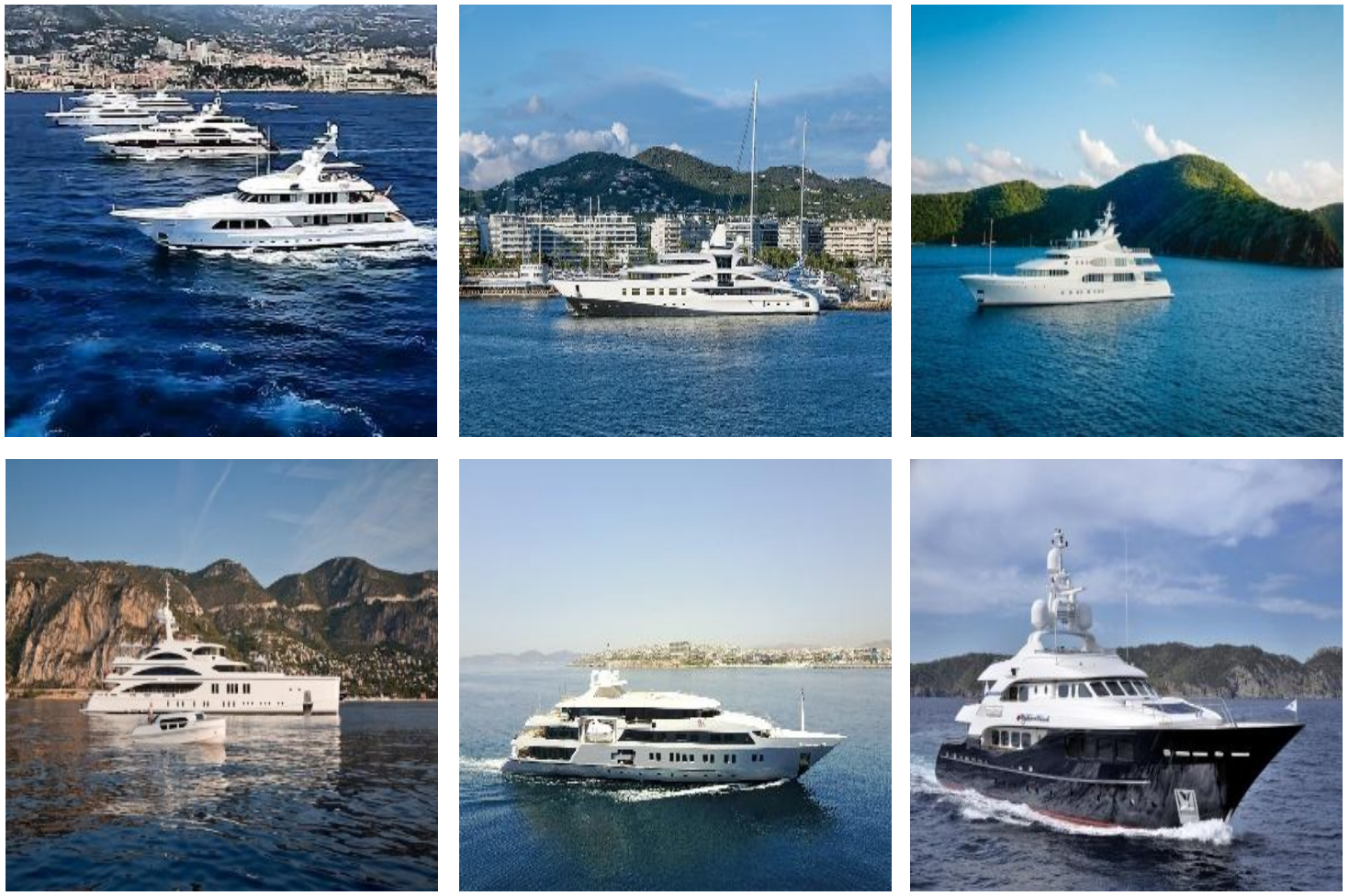}
  \end{subfigure}
  \begin{subfigure}{0.31\textwidth}
    \includegraphics[width=0.98\textwidth]{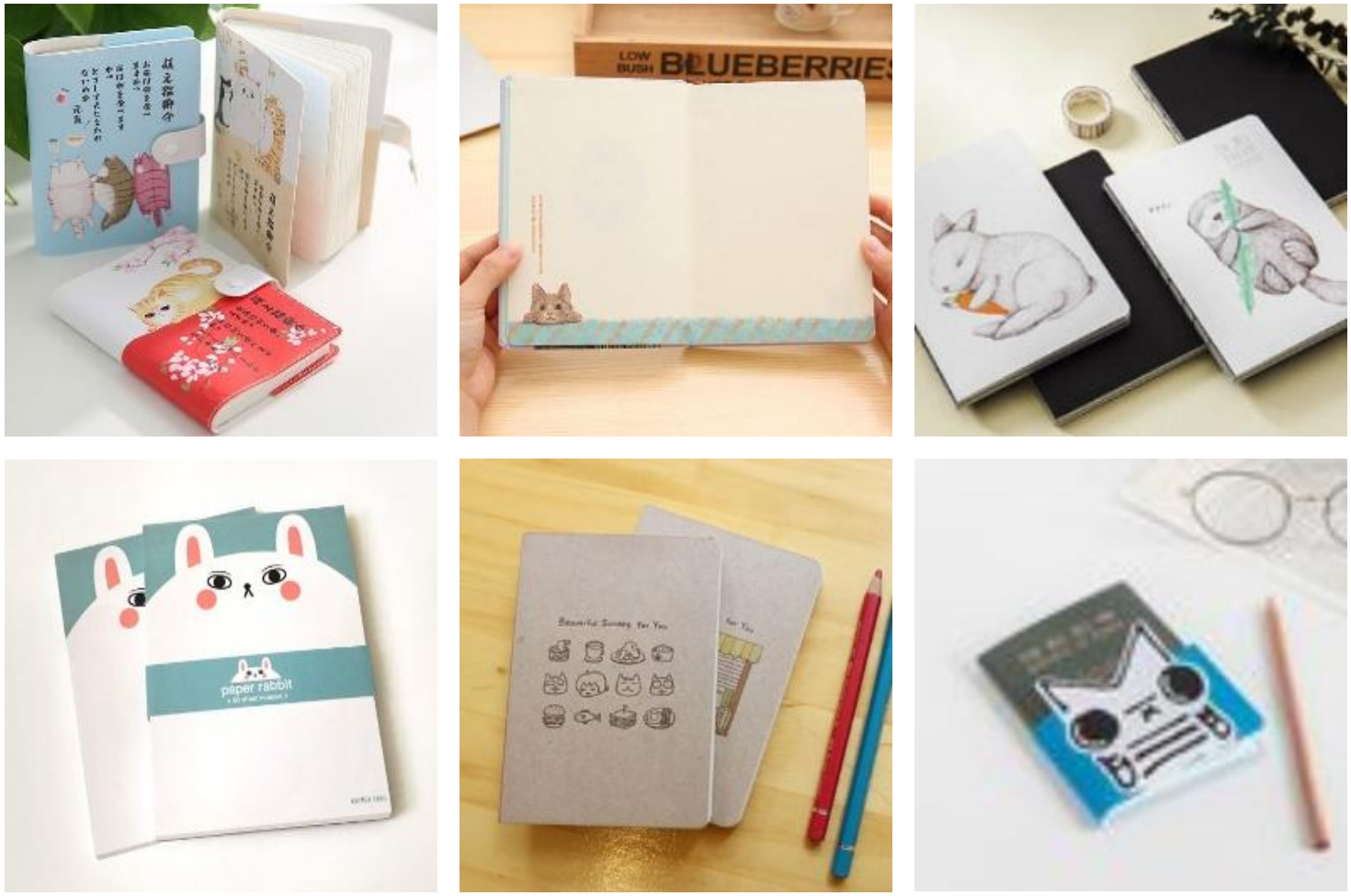}
  \end{subfigure}
  \begin{subfigure}{0.31\textwidth}
    \includegraphics[width=0.98\textwidth]{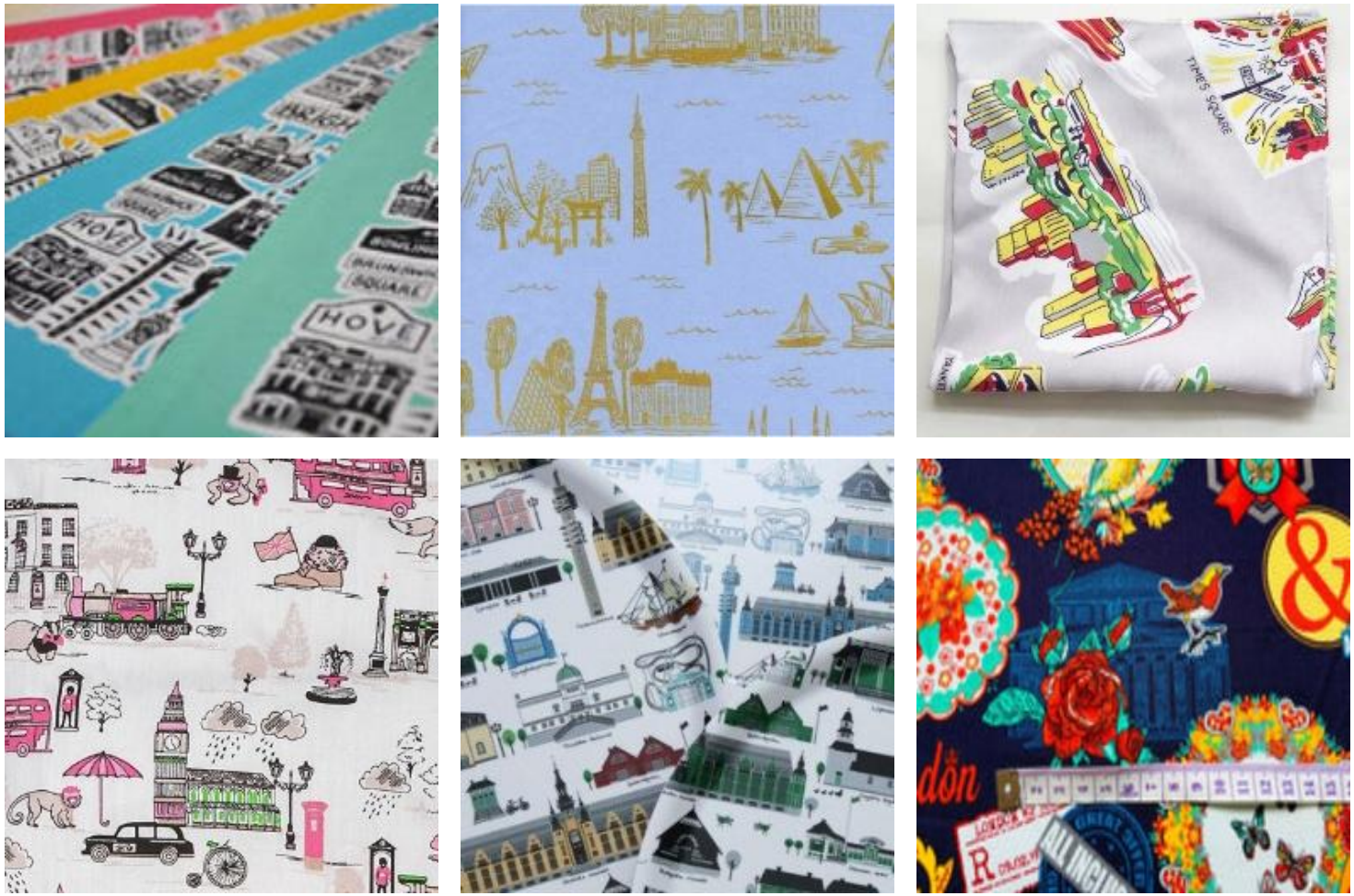}
  \end{subfigure}

  \caption{Visualization of clusters. We randomly present six clusters from the clustering results of the LAION225M dataset.}
\label{vis_cluster}
\vspace{-3mm}
\end{figure*}

\subsection{Cluster Visualization}
The performance of our proposed cluster-instance fusion distillation method is directly influenced by the quality of clustering. Therefore, we randomly visualize six clusters from our 1M clusters. As illustrated in Fig.~\ref{vis_cluster}, the images within each cluster demonstrate a high degree of semantic similarity and exhibit a notable level of purity.

\section{Acknowledgment}
We would like to thank Bin Qin, Lan Wu, and Yuling Wu for their help with the organization of all the datasets. 

\end{document}